\documentclass[10pt,twocolumn,letterpaper]{article}

\usepackage{cvpr}
\usepackage{times}
\usepackage{epsfig}
\usepackage{graphicx}
\usepackage{amsmath}
\usepackage{amssymb}
\usepackage{romannum}
\usepackage{rotating}
\usepackage{multirow}
\usepackage{subcaption}
\usepackage{enumerate}
\usepackage{mathtools}
\DeclarePairedDelimiter{\abs}{\lvert}{\rvert}
\graphicspath{{fg/}}
\usepackage{authblk}
\usepackage{blindtext}
\makeatletter
\renewcommand\AB@affilsepx{, \protect\Affilfont}
\makeatother

\usepackage[breaklinks=true,bookmarks=false]{hyperref}

\cvprfinalcopy 

\def\cvprPaperID{****} 

\ifcvprfinal\fi
\begin{document}

\title{Understanding and Visualizing Deep Visual Saliency Models}
\author[1]{Sen He}
\author[2]{Hamed R. Tavakoli}
\author[3]{Ali Borji}
\author[1]{Yang Mi}
\author[1]{Nicolas Pugeault}
\affil[1]{University of Exeter}
\affil[2]{Aalto University}
\affil[3]{MarkableAI}

\maketitle

\begin{abstract}
Recently, data-driven deep saliency models have achieved high performance and have outperformed classical saliency models, as demonstrated by results on datasets such as the MIT300 and SALICON. Yet, there remains a large gap between the performance of these models and the inter-human baseline. Some outstanding questions include what have these models learned, how and where they fail, and how they can be improved. 
This article attempts to answer these questions by analyzing the representations learned by individual neurons located at the intermediate layers of deep saliency models. 
To this end, we follow the steps of existing deep saliency models, that is borrowing a pre-trained model of object recognition to encode the visual features and learning a decoder to infer the saliency.
We consider two cases when the encoder is used as a fixed feature extractor and when it is fine-tuned, and compare the inner representations of the network.
To study how the learned representations depend on the task, we fine-tune the same network using the same image set but for two different tasks: saliency prediction versus scene classification.
Our analyses reveal that: 1) some visual salient regions (\eg head, text, symbol, vehicle) are already encoded within various layers of the network pre-trained for object recognition, 2) using modern datasets, we find that fine-tuning pre-trained models for saliency prediction makes them favor some categories (\eg head) over some others (\eg text), 3) although deep models of saliency outperform classical models on natural images, the converse is true for synthetic stimuli (\eg pop-out search arrays), an evidence of significant difference between human and data-driven saliency models, and 4) we confirm that, after-fine tuning, the change in inner-representations is mostly due to the task and not the domain shift in the data.
\end{abstract}

\section{Introduction}
The human visual system routinely handles vast amounts of information at about $10^{8}$ to $10^{9}$ bits per second~\cite{borji2018saliency, borji2013state,borji2013quantitative,itti1998model}. An essential mechanism that allows the human visual system to process this amount of information in real time is its ability to selectively focus attention on salient parts of a scene. 
Which parts of a scene and what individual patterns particularly attract the viewer's eyes (\eg salient areas) have been the subject of psychological research for decades, and designing computational models for predicting salient areas is a longstanding problem in computer vision. In recent years, we have observed a surge in the development of data-driven models of saliency based on deep neural networks. Such deep models have demonstrated significant performance improvements in comparison to classical models, which are based on hand-crafted features or psychological assumptions, outperforming them on most benchmarks. 
However, while there still remains a relatively large gap between deep models and the human visual system (see Table~\ref{tab:model}), the performance of deep models appears to have reached a ceiling. This raises the question of what is learned by deep models that drives their superior performance over classical models, and what are the remaining and missing ingredients to attain human-like performance. 
Internal representations of deep object recognition models have been visualized and analyzed extensively in recent years. Such efforts, however, are missing for saliency models and it is unclear how saliency models do what they do.


 \begin{table}[t]
\setlength{\tabcolsep}{2.8pt} 
\renewcommand{\arraystretch}{1} 
\centering
\small
\caption{Five state-of-the-art deep saliency models and their \textit{NSS} scores~\cite{bylinskii2018different} over the MIT300 saliency benchmark~\cite{mit-saliency-benchmark}.} 
\begin{tabular}{|c || c| c|c|}
\hline 
\textbf{}  & \textbf{}   & \textbf{Fine}    &\textbf{} \\
\textbf{Model}  & \textbf{Backbone}   & \textbf{tuning}    &\textbf{NSS} \\

\hline
\hline

Deep gaze \Romannum{2}~\cite{kummerer2017understanding} & VGG-19~\cite{simonyan2014very}      & ${\times}$   & 2.34\\
SAM ~\cite{cornia2018predicting}   &ResNet-50~\cite{he2016deep}/VGG-16    &$\surd$ & 2.34/2.30\\
Deepfix~\cite{kruthiventi2017deepfix} &VGG-16~\cite{simonyan2014very}   &$\surd$  &2.26\\
SALICON~\cite{huang2015salicon} &VGG-16 &$\surd$ &2.12\\
PDP~\cite{jetley2016end}  &VGG-16  &$\surd$ &2.05\\
\hline
Human IO & -  & - &3.29\\
\hline
\end{tabular}
\label{tab:model}
\vspace{-5pt}
\end{table}

In this work, we shed light on what is learned by deep saliency models by analyzing their internal representations. Our contribution are as follows: \\
\indent $\bullet$ We annotate 3 datasets for analyzing the relationship between the deep model's inner representation and the visual saliency in the image. \\
\indent $\bullet$ A new dataset based on synthetic pop-out search arrays is proposed to compare deep and classical saliency models. \\
\indent $\bullet$ We investigate what and how saliency information is encoded in a pre-trained deep model and look into the effect of fine-tuning on inner-representations of the deep saliency models.\\
\indent $\bullet$ Finally, we study the effect of the task type on the inner representations of a deep model by comparing a model fine-tuned for saliency prediction with a model fine-tuned for scene recognition.

\section{Related Work}


\subsection{Deep saliency models}

The SALICON challenge~\cite{jiang2015salicon}, by offering the first large scale dataset for saliency, facilitated the development of deep saliency models. Several such models learn a mapping from deep feature space to the saliency space, where a pre-trained object recognition network acts as the feature encoder. The encoder is then fine-tuned for the saliency task. 
For example, DeepNet~\cite{pan2016shallow} learns saliency using 8 convolutional layers, where only the first 3 layers are initialized from a pre-trained image classification model. 
PDP~\cite{jetley2016end} treats the saliency map as a small scale probability map, and investigates different loss functions for gaze prediction. They also suggested the use of Bhattacharyya distance when the gaze map is treated as a small scale probability map. 
The SALICON~\cite{huang2015salicon} model uses multi-resolution inputs, and combines feature representations in the deep layers for saliency prediction. 
Deepfix~\cite{kruthiventi2017deepfix} combines deep architectures of VGG, GoogleNet~\cite{szegedy2015going}, and Dilated convolutions~\cite{yu2015multi} in a network and adds a central bias, to achieve a higher performance than previous models. SalGAN~\cite{pan2017salgan} uses an encoder-decoder architecture and proposes the binary cross entropy (BCE) loss function to perform pixel-wise (rather than image-wise) saliency estimation. After pre-training the encoder-decoder, it uses a Generative Adversarial Network (GAN)~\cite{goodfellow2014generative} to boost performance. 
DVA~\cite{wang2018deep} uses multiple layer's representations, builds a decoder for each layer, and fuses them at the final stage for pixel-wise gaze prediction. 
SAM~\cite{cornia2018predicting} uses an attention module and a LSTM~\cite{hochreiter1997long} network to attend to different salient regions in the image.
DeepGaze \Romannum{2}~\cite{kummerer2017understanding} uses the features at different layers of a pre-trained deep model and combines them with the prior knowledge (\ie center-bias).
DSCLRCN~\cite{liu2018deep} uses multiple inputs by adding a contextual information stream, and concatenates the original representation and the contextual representation into a LSTM network for the final prediction.

\subsection{Visualizing deep neural networks}
The success of deep convolutional neural networks has raised the question of what representations are learned by neurons located in intermediate and deep layers. One approach towards understanding how CNNs work and learn is to visualize individual neurons' activations and receptive fields. 
Zeiler and Fergus~\cite{zeiler2014visualizing} proposed a deconvolution network in order to visualize the original patterns that activate the corresponding activation maps. A deconvolution network consists of the three steps of unpooling, transposed convolution, and the ReLU operation. 
Yosinski \etal~\cite{yosinski2015understanding} developed two tools for understanding deep convolutional neural networks. The first of these tools is designed to visualize the activation maps at different layers for a given input image. The second tool aims to estimate the input pattern which a network is maximally attuned to for a given object class. In practice, the last layer of a deep neural network typically consists of one neuron per object class. Yosinski \etal proposed to use gradient ascent (with regularization) to find the input image that maximizes the output of a specific neuron in regard to a specific object class. Hence, they derive the optimum input that appeals to the network for a specific class.

Both visualization methods discussed above are essentially qualitative. In contrast, Bau \etal~\cite{netdissect2017} proposed a quantitative method to give each activation map a \textit{semantic meaning}. 
In their work, they proposed a dataset with 6 image categories and 63,305 images for network dissection, where each image is labeled with pixel-wise semantic meaning. At first, they forward all images in the dataset into a pre-trained deep model. For each activation map inside the model, different inputs have different patterns. Then, they compute the distribution of each unit activation map over the whole dataset, and determine a threshold for each unit based on its activation distribution. With the threshold for each unit, the activation map for each input image is quantized to a binary map. Finally, they compute the intersection over union (IOU) between the quantized activation map and the labeled ground truth to determine what objects or object parts a unit is detecting. 

The aforementioned approaches provide useful insight into the internals of deep neural networks trained on ImageNet for the classification task. However, our understanding of the internal representations of deep saliency prediction models is somewhat limited. Bylinskii \etal~\cite{bylinskii2016should} tried to understand deep models for saliency prediction. But their study was mostly focused on where models fails, rather than how they compute saliency. To our best of knowledge, our work is the first to study the representations learned by deep saliency models\footnote{All the codes, data, models and other details in the paper can be found at \underline{\url{https://github.com/SenHe/uavdvsm}}}.

\section{Data and Annotation}
We first introduce the data used in our experiments as well as our proposed annotated dataset. 

\textbf{SALICON}: SALICON~\cite{jiang2015salicon} is the largest database for saliency prediction at the moment. It contains $10,000$ training images, $5,000$ validation images and $5,000$ testing images. Here, we use it to fine-tune a pre-trained model for saliency prediction.

\begin{figure}[t]
\centering
\begin{subfigure}[b]{0.32\linewidth}
\includegraphics[width=\linewidth]{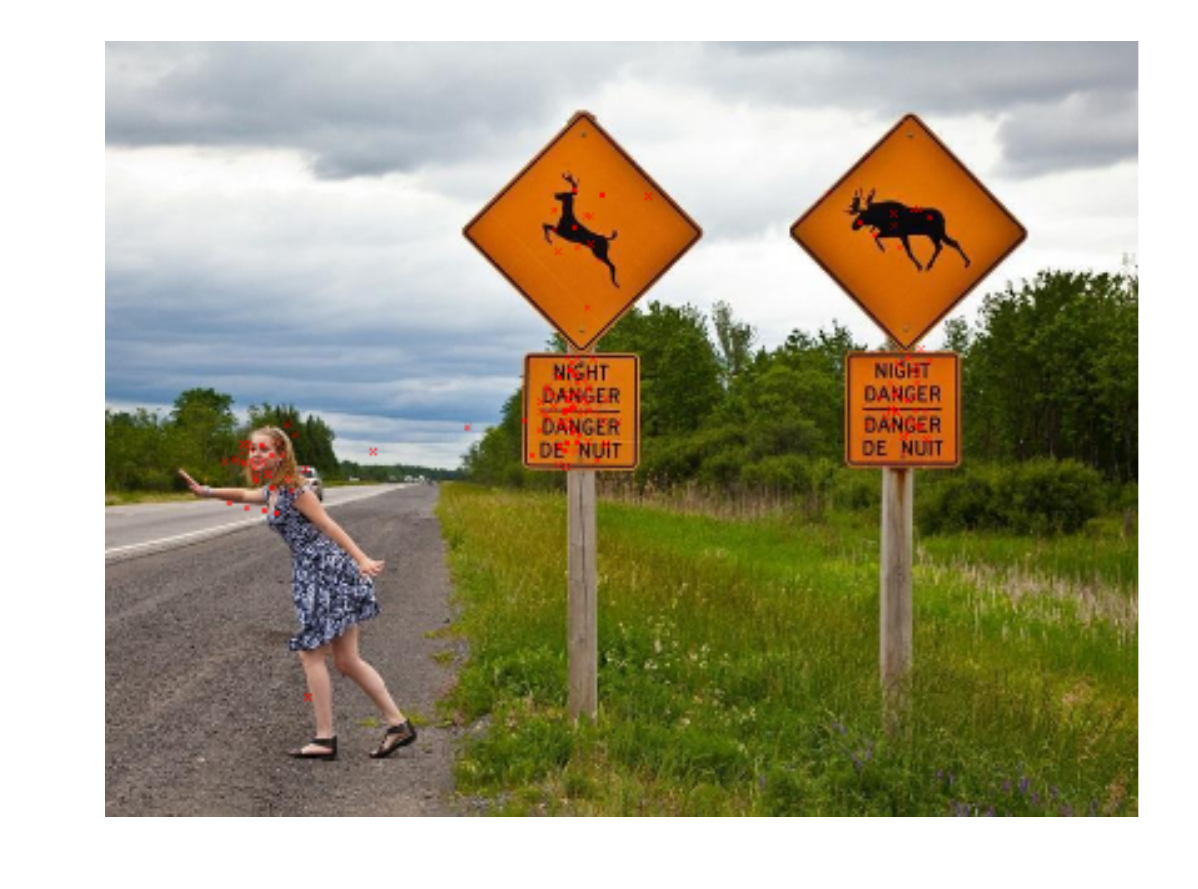}
\vspace{-20pt}
\caption{\textit{image}}
\end{subfigure}
\begin{subfigure}[b]{0.32\linewidth}
\includegraphics[width=\linewidth]{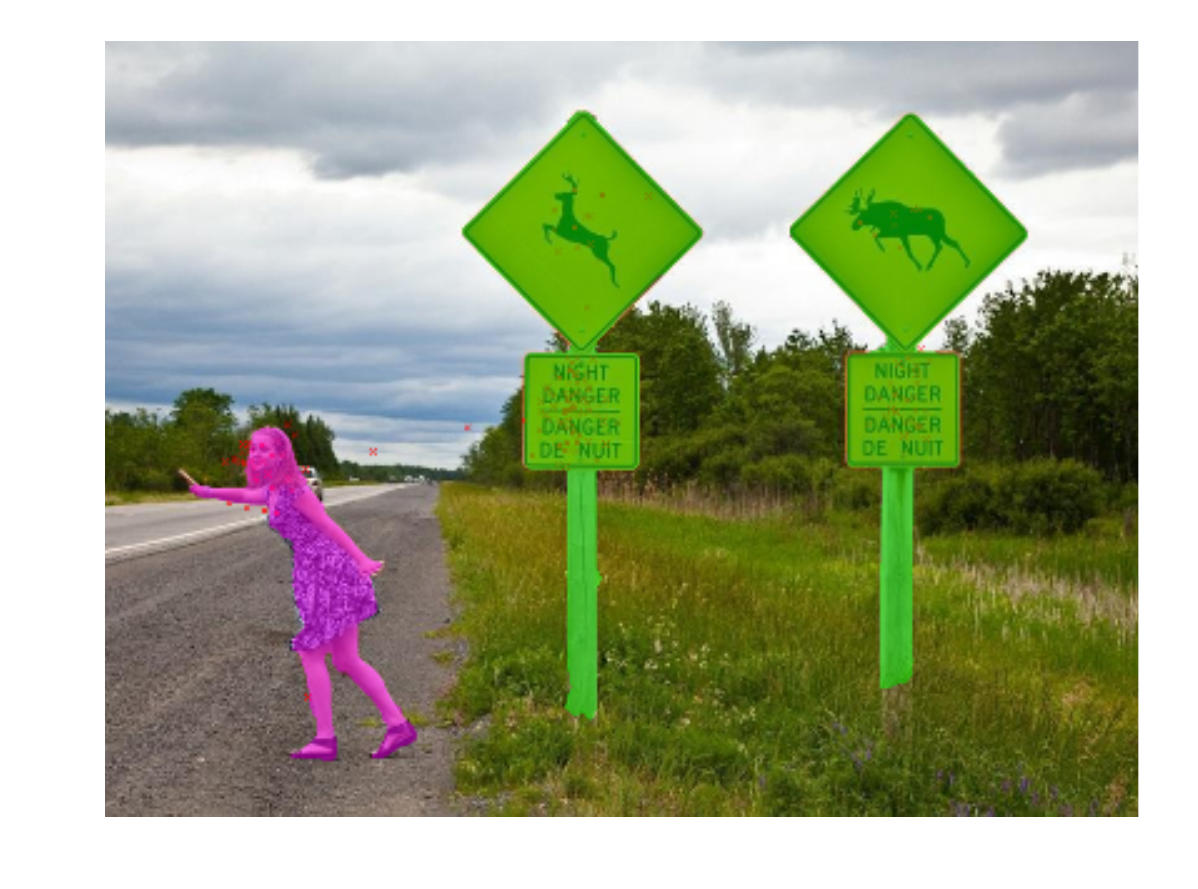}
\vspace{-20pt}
\caption{\textit{OSIE}}
\end{subfigure}
\begin{subfigure}[b]{0.32\linewidth}
\includegraphics[width=\linewidth]{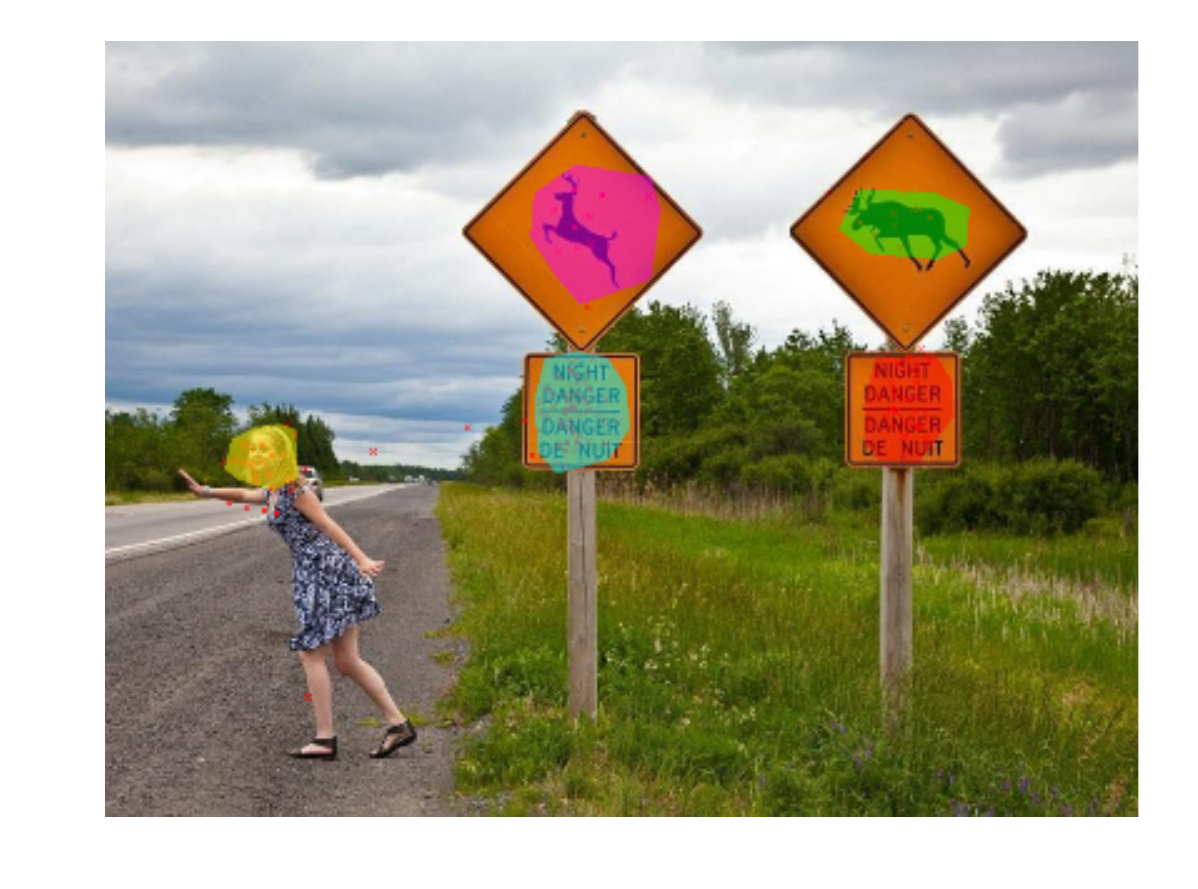}
\vspace{-20pt}
\caption{\textit{OSIE-SR}}
\end{subfigure}
\caption{(a) An example image from the OSIE dataset, (b) OSIE annotation, and (c) the re-annotated OSIE-SR labels.}
\label{fig:anno}
\end{figure}

\textbf{OSIE-SR}: This acronym stands for the ``OSIE Saliency Re-annotated" (\ie annotations of salient regions). The original OSIE dataset~\cite{xu2014predicting} has 700 images with rich semantics. It contains eye movements of 15 subjects for each image recorded during free-viewing, and also has the annotated masks for objects in the image according to 12 attributes. For our analysis, we extract clusters of fixation locations, called \textit{salient regions}.  
We, then, manually annotate each salient region as belonging to one of the 12 saliency categories, including: person head, person part, animal head, animal part, object, text, symbol, vehicle, food, drink, plant, and other. Similar categories have been exploited in previous research (\eg ~\cite{bylinskii2016should,xu2014predicting}). Fig.~\ref{fig:anno} provides an example where each annotated region has a label according to its salient category. The re-annotated data is used to measure the association between inner representations (activation maps) in the deep model (pre-trained and fine-tuned) and each salient category.

\begin{figure}[t]
\centering
\begin{subfigure}[b]{0.23\linewidth}
\includegraphics[width=\linewidth]{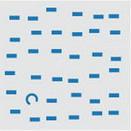}
\end{subfigure}
\begin{subfigure}[b]{0.23\linewidth}
\includegraphics[width=\linewidth]{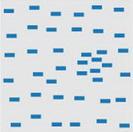}
\end{subfigure}
\begin{subfigure}[b]{0.23\linewidth}
\includegraphics[width=\linewidth]{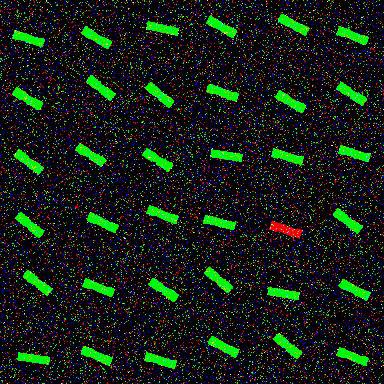}
\end{subfigure}
\begin{subfigure}[b]{0.23\linewidth}
\includegraphics[width=\linewidth]{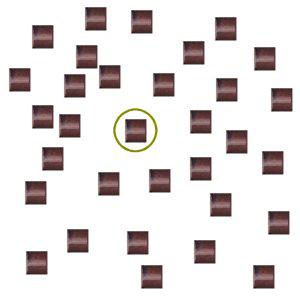}
\end{subfigure}
\caption{Example synthetic pop out search arrays} 
\label{fig:syn}
\end{figure}

\textbf{Synthetic Images}: We selected 80 synthetic search arrays~\cite{healey2012attention}, often used in pop-out experiments (Fig.~\ref{fig:syn}). 
This dataset contains various pop-out patterns where a target stands out from the rest of the items in terms of color, orientation, density, curvature, etc. We provide mask annotation for the salient (\ie pop-out) region in each image. This database is used to compare deep models and classical models on their ability to detect targets that pop-out in simple scenes. We also use it to study inner representations of deep models over synthetic patterns.

\textbf{SALICON-SAL-SCE}: We select a subset of images from the SALICON dataset and annotate each image with a scene category label based on the categories of the Place-CNN dataset~\cite{zhou2017places}. We removed images belonging to the scene categories with fewer than 50 images. Eventually, we are left with 6,107 images with both fixation maps and scene category labels (26 categories in total, see supplement). We use this database to compare the effect of saliency prediction and scene recognition on learned inner representations.

\section{Methods}

\subsection{Modified NSS score}
We propose using normalized scanpath score (NSS) within salient regions as a method to interpret the inner representations. We, thus, can look into the association between the activation maps and the salient regions of the image for analyzing the inner representations of the deep visual saliency models. To implement this, we first forward each image into the deep model and extract the activation maps from different layers in the model. Then, the association between each activation map (\textit{actm}) and each salient region ({\textit{sr}}) in the image is computed as:
\begin{align}
\small
Assoc(actm_{ij},sr_{lk}) = NSS(actm_{ij},fix_l \boldsymbol{\cdot} mask_{lk}) \label{eq:1}
\end{align}
\noindent where $actm_{ij}$ is the $j_{th}$ activation map from the $i_{th}$ layer for the input $l_{th}$ image, and $sr_{lk}$ is the $k_{th}$ salient region in the $l_{th}$ image. $fix_l$ is the fixation on the $l_{th}$ image, and $mask_{lk}$ is the annotated polygon mask in $l_{th}$ image for region $k$. It is worth noting that all activation maps were normalized and reshaped to the size of the input image.

\subsection{Saliency model}
For our analysis, we develop a saliency model using the convolutional part of VGG-16 (conv1-1 to conv5-3) and a simple $1\times1$ convolutional layer on top of the conv5-3 layer. The model has a single resolution with input of size $224\times224$ and is optimized with $-NSS$ as the loss function. 
We consider 2 setups to analyze the representations. 

{$\bullet$ \bf Setup I:} We first look into the inner representations without fine-tuning the VGG part, \ie we \emph{only} learn the last $1\times1$ convolution layer, motivated by the performance of some of the existing models that achieve state-of-the-art performance without fine-tuning. This can be observed in Table \ref{tab:model}. In other words, we analyze what types of saliency information exist in the pre-trained VGG model and how it is distributed within different layers. In this setup, we analyze conv4-1 to conv5-3 layers, which correspond to the last two blocks in VGG-16. The activation maps from layers below conv4-1 are sensitive to edge-like patterns and do not correspond to annotated regions. We, thus, do not include them. 

{$\bullet$ \bf Setup II:} We then look into the fine-tuned model. In this setup, we learn the last $1\times1$ convolution layer and fine-tune the VGG part of the model for different number of layers (each time from scratch) and examine the responses of neurons in the conv5-3 layer.

\subsection{Local saliency statistics}
In the OSIE-SR dataset, each region corresponds to one saliency category (\ie has one label). To compute the statistics for saliency category $c$ in each layer, we compute the mean value of the top 10 activation maps with high mean NSS values in Eq. \eqref{eq:1} for all the regions of category $c$. We also compute the number of activation maps whose mean NSS value is above a threshold (T) in Eq. \eqref{eq:1} for all regions for category $c$. 

\section{Analysis of Learned Representations}
How does fine-tuning the VGG neurons for saliency prediction affect the network inner representation? To answer this question, we train two saliency models, one keeping the CNN features fixed during the training and the other one fine-tuning the CNN features in conjunction with the $1\times1$ convolution layer for
saliency prediction (corresponding to the two setups mentioned above).

\begin{table}[t!]
\setlength{\tabcolsep}{1.6pt} 
\renewcommand{\arraystretch}{1.1} 
\caption{Inner representations in the pre-trained visual saliency model (setup I; \ie training the $1\times1$ convolution layer) at different layers for all types of saliency categories (Please see text for details).}
\centering
\footnotesize
\begin{tabular}{|c||c|c|c|c|c|c|c|c|c|c|c|c|}
\hline
\multirow{5}{*}{ \rotatebox[origin=c]{90}{ \textit{layer}}}
&
\multirow{5}{*}{ \rotatebox[origin=c]{90}{ \textit{person head}}}&
\multirow{5}{*}{ \rotatebox[origin=c]{90}{ \textit{person part}} }&
\multirow{5}{*}{\rotatebox[origin=c]{90}{ \textit{animal head}} }&
\multirow{5}{*}{\rotatebox[origin=c]{90}{ \textit{animal part}} }&
\multirow{5}{*}{\rotatebox[origin=c]{90}{ \textit{object}} }&
\multirow{5}{*}{\rotatebox[origin=c]{90}{ \textit{text}}}&
\multirow{5}{*}{\rotatebox[origin=c]{90}{ \textit{symbol}}}&
\multirow{5}{*}{\rotatebox[origin=c]{90}{ \textit{vehicle}}} &
\multirow{5}{*}{\rotatebox[origin=c]{90}{ \textit{food}}} &
\multirow{5}{*}{\rotatebox[origin=c]{90}{ \textit{plant}}} &
\multirow{5}{*}{\rotatebox[origin=c]{90}{ \textit{drink}}}&
\multirow{5}{*}{\rotatebox[origin=c]{90}{ \textit{other}}} \\
& & & & & & & & & & & &\\
& & & & & & & & & & & &\\
& & & & & & & & & & & &\\
& & & & & & & & & & & &\\
\hline
\hline
&\multicolumn{12}{c|}{mean NSS for top 10 activation maps}\\
\hline
conv5-3  &1.21 &0.98 &2.04 &1.6 &0.69 &1.1 &1.07 &1.53 &1.25 &1.21 &1.48 &0.57    \\ 
conv5-2  &2.57&1.64 &2.89 &\textbf{1.91} &1.07 &1.59 &1.91 &2.18 &1.59 &\textbf{2.05} &\textbf{2.13} & 0.92   \\
conv5-1  &2.95 &1.58 &2.6 &1.67 &1.16 &1.91 &2.06 &2.48 &\textbf{1.8} &1.96 &1.85 &\textbf{1}    \\
conv4-3 &\textbf{3.46} & 1.67 &\textbf{2.93} &1.5&\textbf{1.27}&\textbf{2.4}&\textbf{2.37}&\textbf{2.71}&1.54&1.62&1.88&0.91\\
conv4-2 &2.85& \textbf{1.78}& 2.63&1.39& 1.19&2.04&2.05&2.33&1.48&1.59&1.53&0.74\\
conv4-1 &2.08&1.56& 1.89&1.18&1.11&2.17&1.99&1.93&1.38&1.58&1.33&0.72\\
\hline
\hline
&\multicolumn{12}{c|}{\# activation maps above threshold ($T=1.5$)}\\
\hline
conv5-3  &2 &1 &21 &7 &0 &2 &1 &4 &1 &2 &4 &0    \\ 
conv5-2  &\textbf{41}&11 &\textbf{50} &\textbf{21} &0 &5 &19 &\textbf{33} &8 &\textbf{12} &15 & 0   \\
conv5-1  &27 &7 &35 &7 &0 &9 &16 &30 &\textbf{9} &11 &\textbf{23} &0    \\
conv4-3 &35 &9 &40 &4& \textbf{1}& 12 & \textbf{20} &30 & 3 & 5 & 13 & 0\\
conv4-2 &35 & \textbf{12} & 18 & 2 & 0 & 11 & 14 & 23 & 4 & 6 & 4 & 0\\
conv4-1 & 23 & 5 & 18 & 0 & 0 & \textbf{18} & 17 & 21 & 3 & 5 & 2 & 0\\
\hline
\end{tabular}
\label{tab:bf}
\end{table}

\begin{figure*}[t!]
\centering
\begin{subfigure}[b]{0.19\linewidth}
\includegraphics[width=\linewidth]{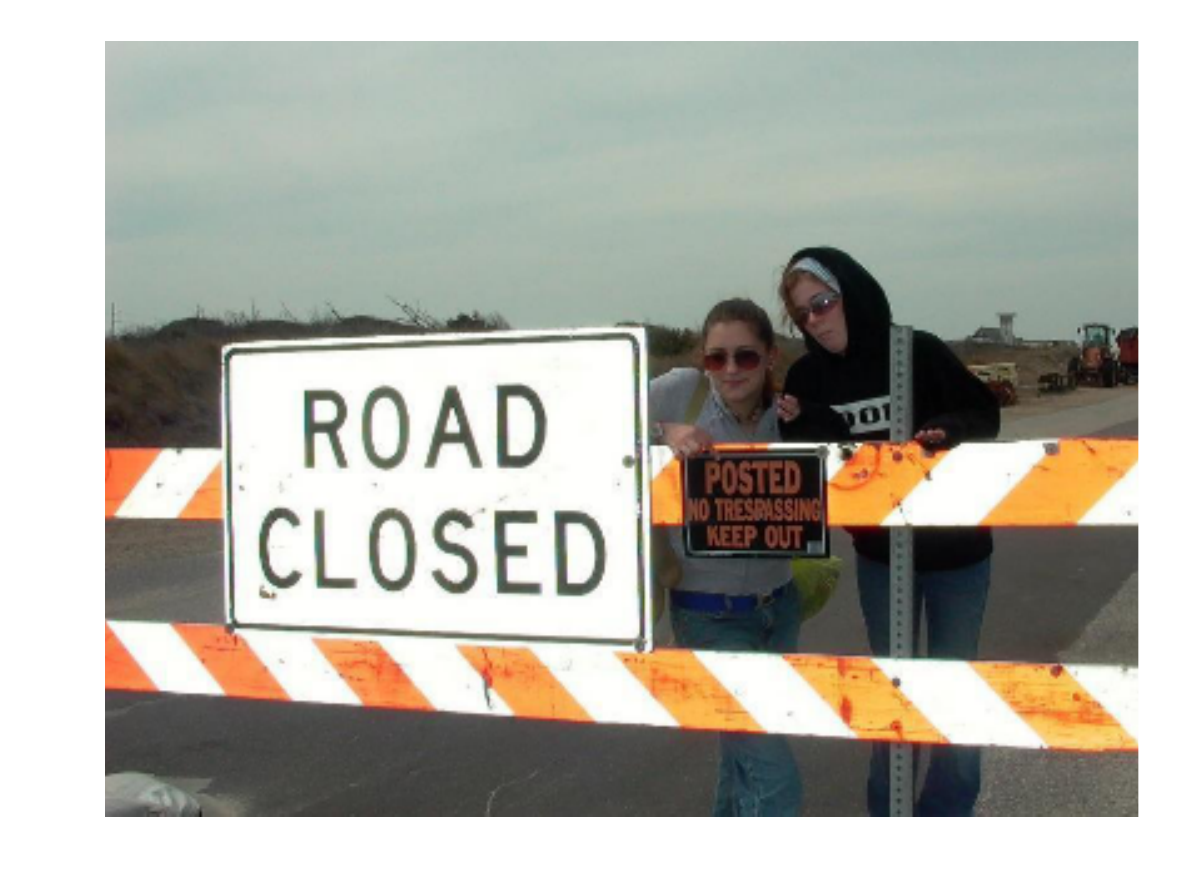}
\end{subfigure}
\begin{subfigure}[b]{0.19\linewidth}
\includegraphics[width=\linewidth]{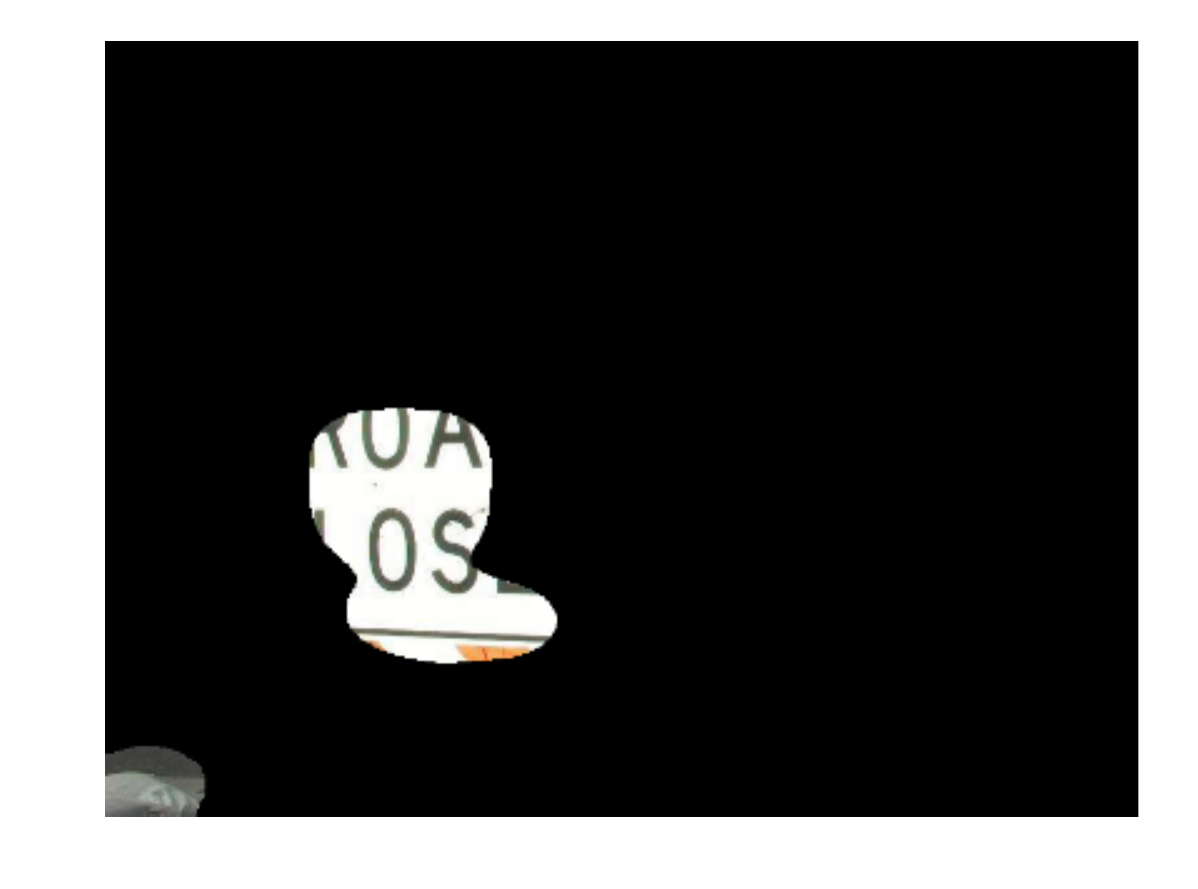}
\end{subfigure}
\begin{subfigure}[b]{0.19\linewidth}
\includegraphics[width=\linewidth]{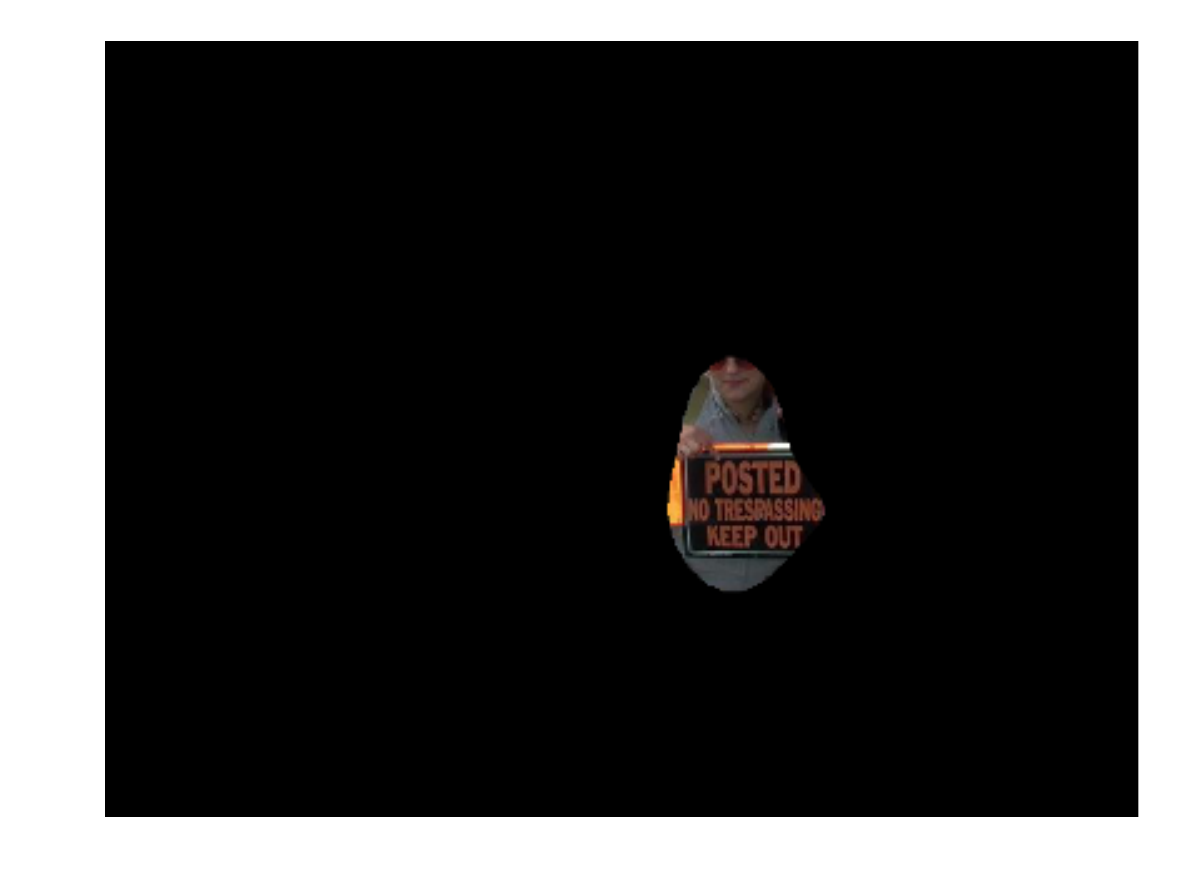}
\end{subfigure}
\begin{subfigure}[b]{0.19\linewidth}
\includegraphics[width=\linewidth]{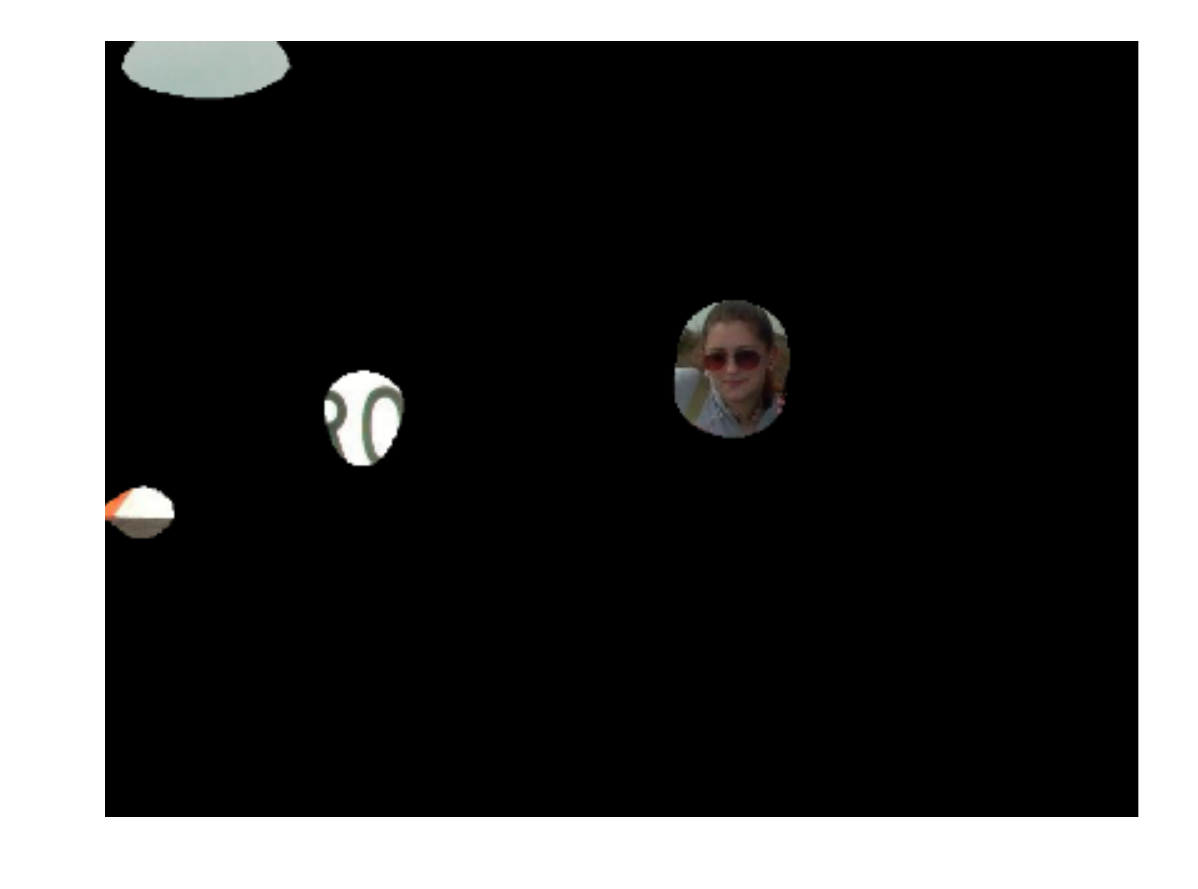}
\end{subfigure}
\begin{subfigure}[b]{0.19\linewidth}
\includegraphics[width=\linewidth]{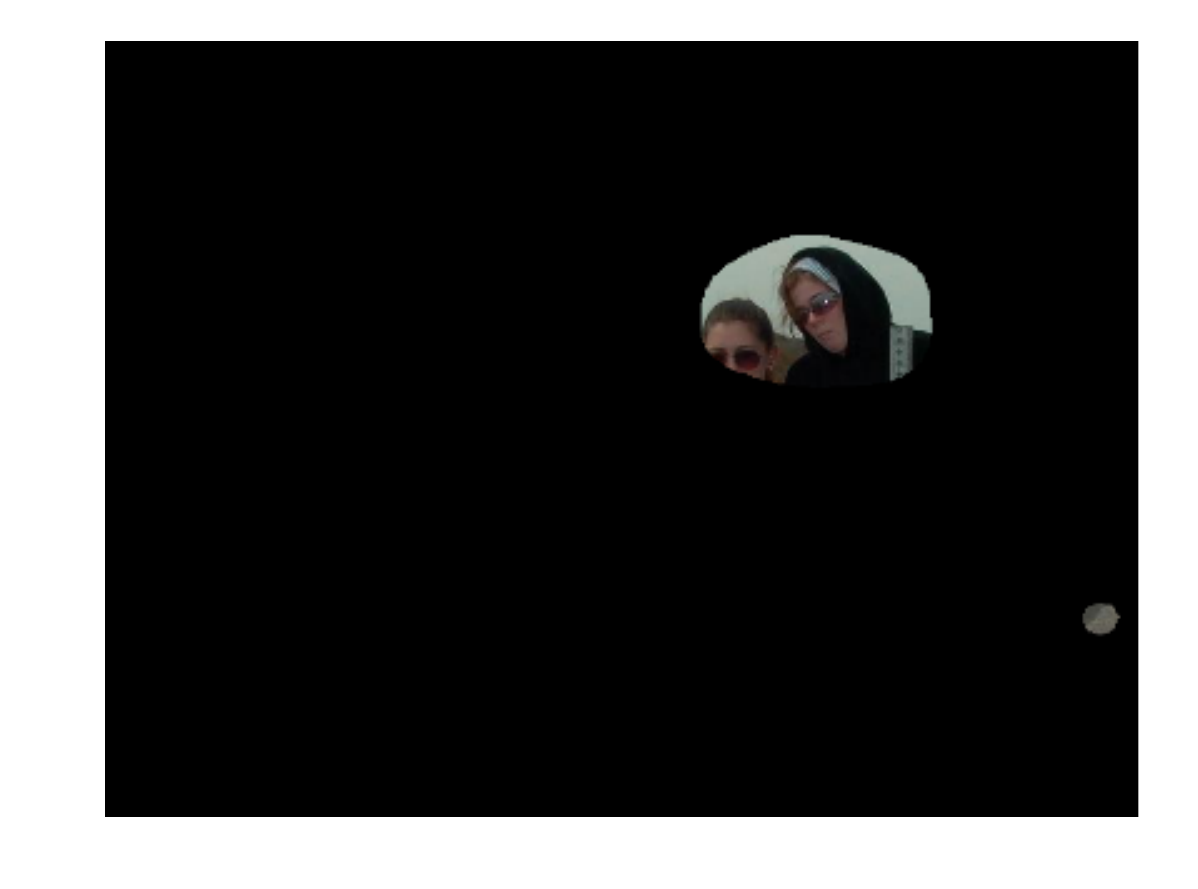}
\end{subfigure}
\begin{subfigure}[b]{0.19\linewidth}
\includegraphics[width=\linewidth]{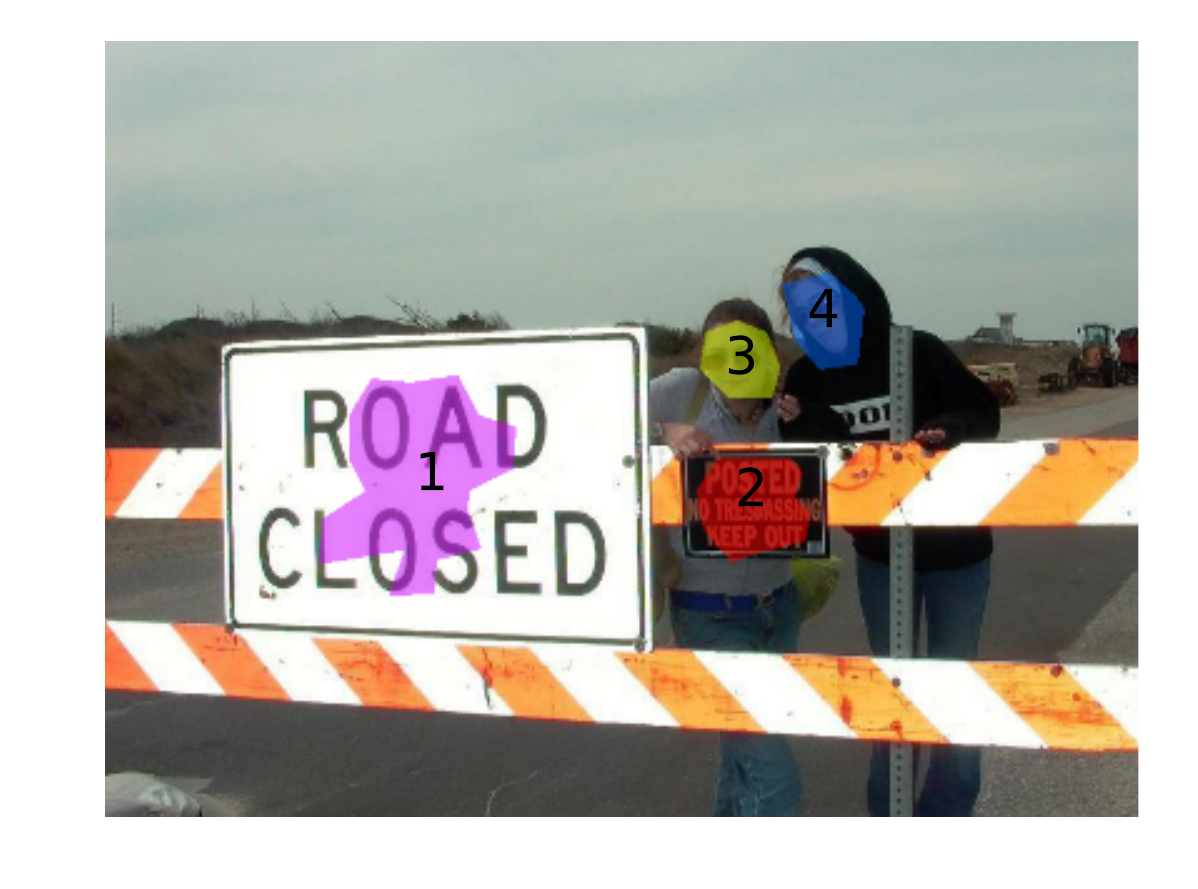}
\end{subfigure}
\begin{subfigure}[b]{0.19\linewidth}
\includegraphics[width=\linewidth]{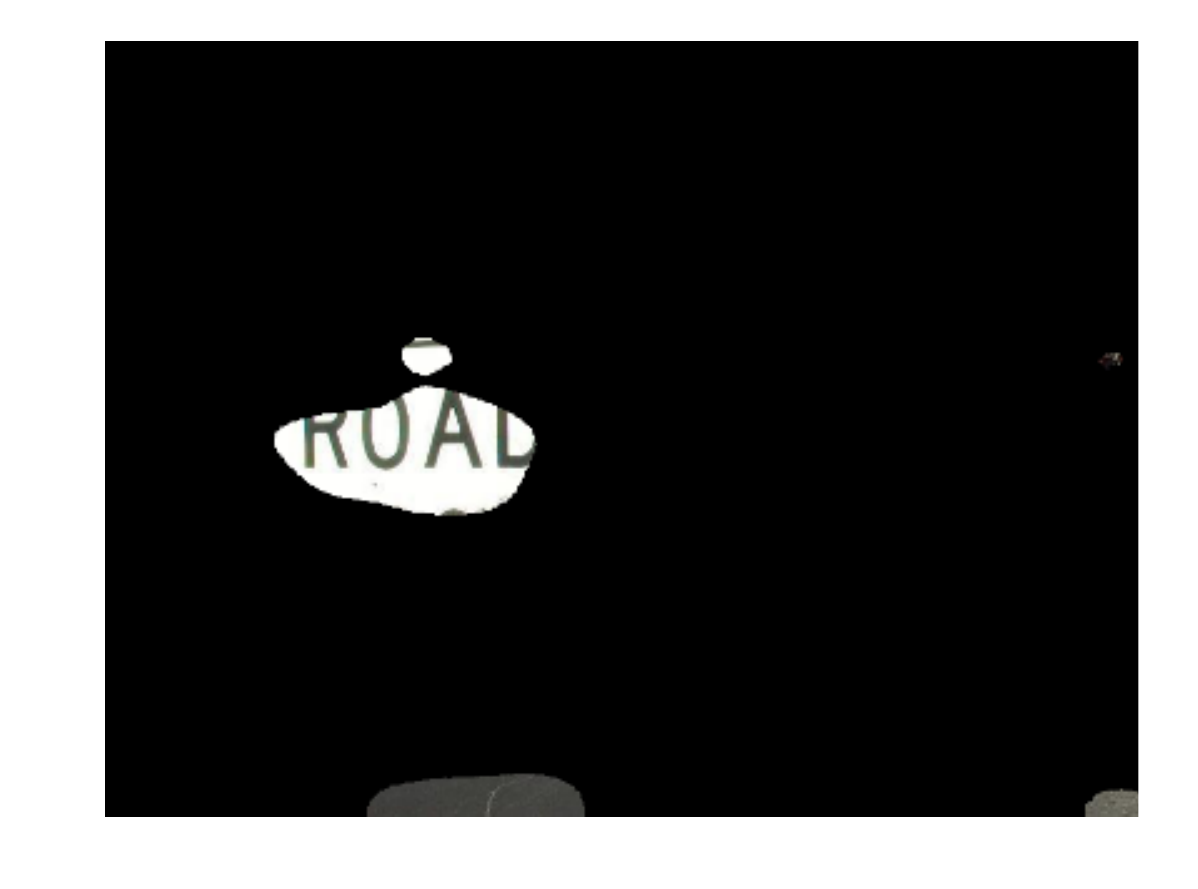}
\end{subfigure}
\begin{subfigure}[b]{0.19\linewidth}
\includegraphics[width=\linewidth]{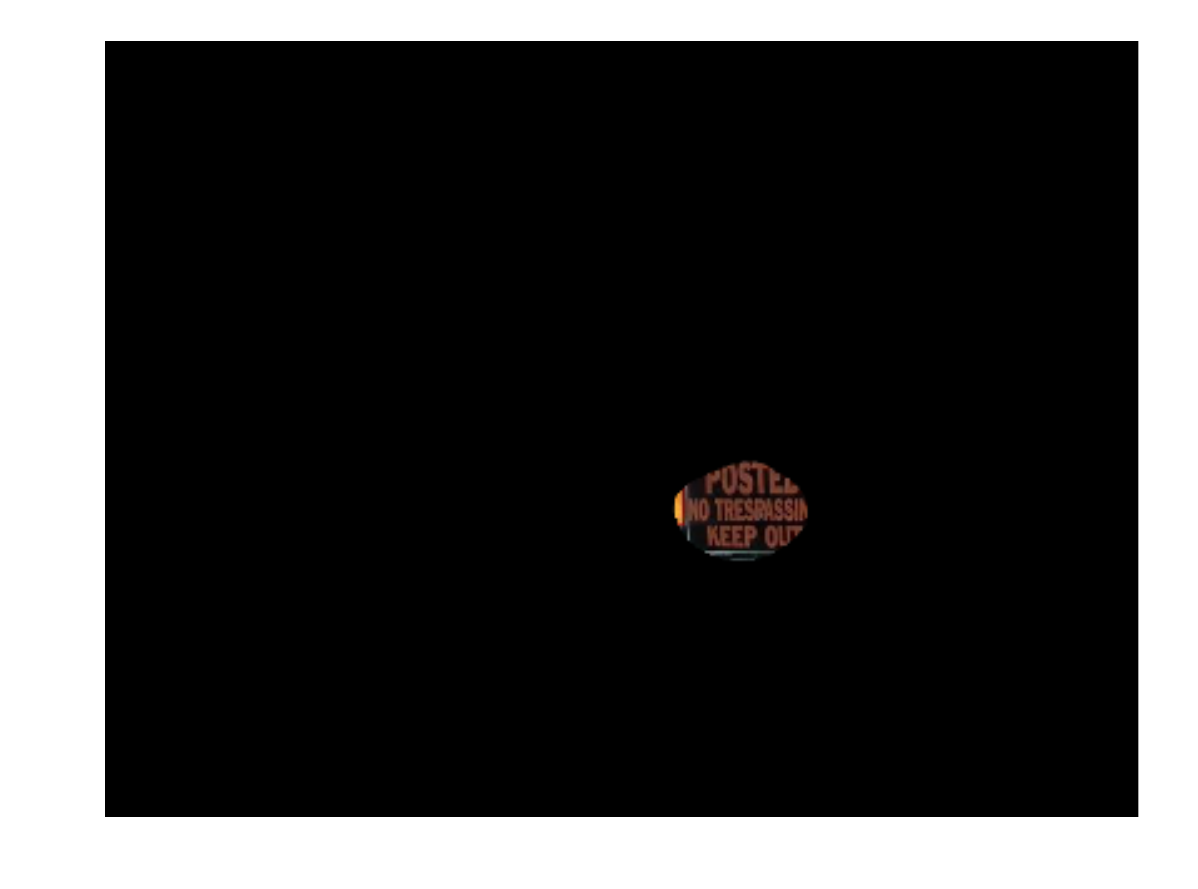}
\end{subfigure}
\begin{subfigure}[b]{0.19\linewidth}
\includegraphics[width=\linewidth]{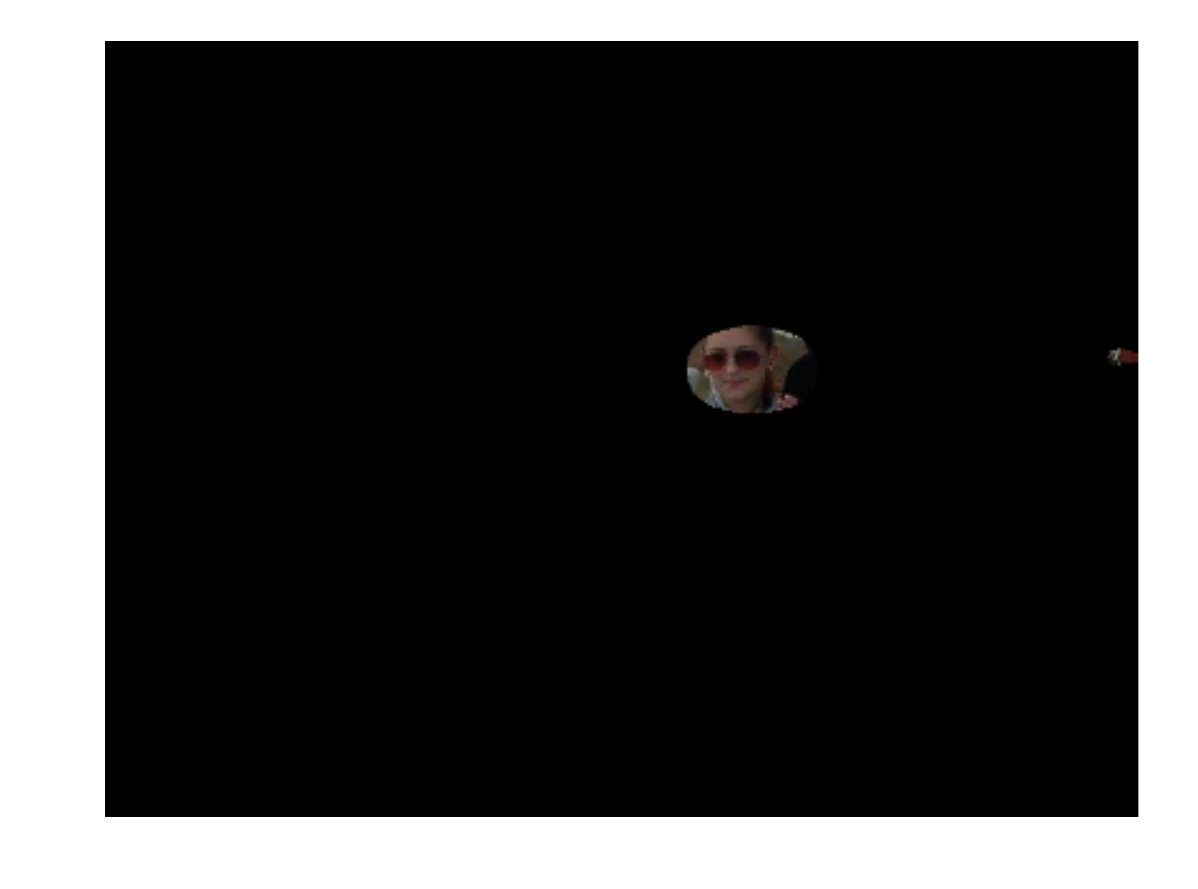}
\end{subfigure}
\begin{subfigure}[b]{0.19\linewidth}
\includegraphics[width=\linewidth]{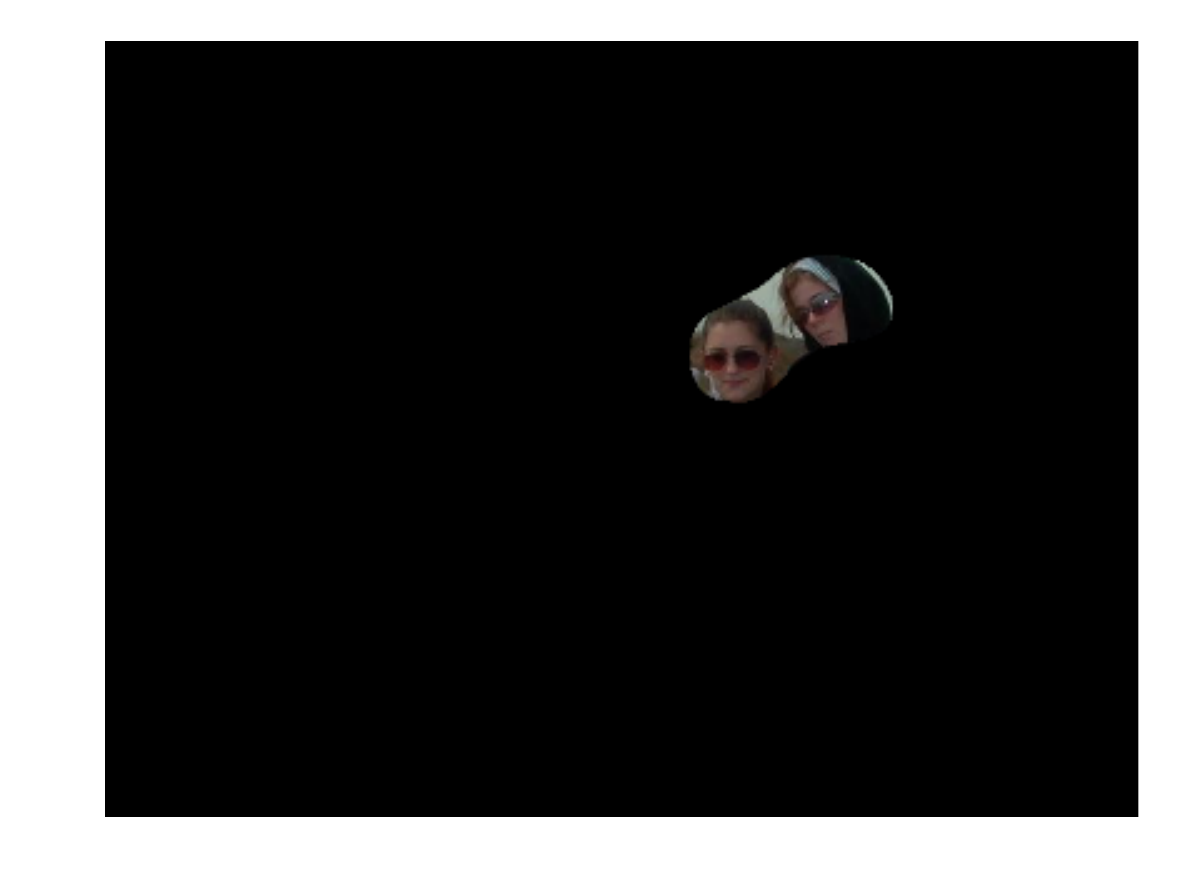}
\end{subfigure}
\caption{An example of the pre-trained model inner representation (Setup I). Top row: the original image, and the activation maps with highest activation at salient regions from layer conv5-1. Bottom row: the image with masked salient regions (1,2,3,4) and the activation maps that best respond to the salient regions of conv4-3.}
\label{fig:pre_exa}
\vspace{-10pt}
\end{figure*}

\subsection{Saliency representation before fine-tuning}
Table~\ref{tab:bf} depicts the statistics of the inner representations within different layers of the deep visual saliency model, where the convolution part has not been fine-tuned (setup I). From Table~\ref{tab:bf}, we can see that many visual saliency categories, including: \textit{person head}, \textit{animal head}, \textit{text}, \textit{symbol}, \textit{vehicle} and \textit{drink}, have been encoded in the pre-trained CNN features. We observe not only high mean NSS scores, but also a large number of active response maps for each saliency category. We also observe that the visual saliency information is encoded within various layers, \eg \textit{person head}, \textit{animal head}, and \textit{text} are present in conv4-3. Fig.~\ref{fig:pre_exa} visualizes some examples of the activation maps in the model without fine-tuning the VGG features. As depicted, there is a relatively high association between salient regions and activation maps.

\begin{table}[t!]
\setlength{\tabcolsep}{2.3pt} 
\renewcommand{\arraystretch}{1.1} 

\caption{
Inner representations in the last convolutional layer (conv5-3) before and after fine-tuning for all types of saliency categories. 0 in number of fine tuned layers indicate setup I, otherwise setup II.
}
\centering
\footnotesize
\begin{tabular}{|c||c|c|c|c|c|c|c|c|c|c|c|c|}
\hline

\multirow{5}{*}{\rotatebox[origin=c]{90}{\# layers tuned}}&
\multirow{5}{*}{\rotatebox[origin=c]{90}{ \textit{person head}}}&
\multirow{5}{*}{\rotatebox[origin=c]{90}{ \textit{person part}} }&
\multirow{5}{*}{\rotatebox[origin=c]{90}{ \textit{animal head}} }&
\multirow{5}{*}{\rotatebox[origin=c]{90}{ \textit{animal part}} }&
\multirow{5}{*}{\rotatebox[origin=c]{90}{ \textit{object}} }&
\multirow{5}{*}{\rotatebox[origin=c]{90}{ \textit{text}}}&
\multirow{5}{*}{\rotatebox[origin=c]{90}{ \textit{symbol}}}&
\multirow{5}{*}{\rotatebox[origin=c]{90}{ \textit{vehicle}}} &
\multirow{5}{*}{\rotatebox[origin=c]{90}{ \textit{food}}} &
\multirow{5}{*}{\rotatebox[origin=c]{90}{ \textit{plant}}} &
\multirow{5}{*}{\rotatebox[origin=c]{90}{ \textit{drink}}}&
\multirow{5}{*}{\rotatebox[origin=c]{90}{ \textit{other}}} \\
& & & & & & & & & & & &\\
& & & & & & & & & & & &\\
& & & & & & & & & & & &\\
& & & & & & & & & & & &\\
\hline
\hline
&\multicolumn{12}{c|}{mean NSS for top 10 activation maps}\\
\hline
0  &1.21 &0.98 &2.04 &1.6 &0.69 &1.1 &1.07 &1.53 &1.25 &1.21 &1.48 &0.57    \\ 
\hline
1  &2.61&1.63 &2.44 &1.7 &\textbf{1.31} &1.17 &1.56 &\textbf{2.21} &1.97 &\textbf{1.89} &\textbf{1.99} & 1.09   \\
2  &3.25 &1.77 &2.75 &1.78 &1.25 &1.32 &1.56 &1.91 &1.89 &1.66 &1.65 &1.12    \\
3  &3.28 &1.78 &3.04 &1.79 &1.28 &1.34 &1.57 &1.94 &1.96 &1.81 &1.68 &\textbf{1.2}    \\
4  &\textbf{3.38} &\textbf{1.83} &\textbf{3.08} &1.78 &1.24 &1.25 &1.47 &1.97 &1.87 &1.73 &1.63 &1.12    \\
5  &3.32 &1.78 &2.84 &\textbf{1.82} &1.28 &\textbf{1.38} &\textbf{1.59} &2.1 &1.95 &1.75 &1.77 &1.17    \\
6  &3.08 &1.83 & 2.57&1.78 &1.25 &1.25 & 1.45&2.03 &\textbf{1.99} &1.8 &1.66 &1.11    \\
\hline
all  &2.85 &1.75 &2.36 &1.65 &1.14 &1.2 & 1.41&1.72 &1.84 &1.71 &1.33 &1.07    \\
\hline
\hline
&\multicolumn{12}{c|}{\# activation maps above threshold ($T=1.5$)}\\
\hline
0  &2 &1 &21 &7 &0 &\textbf{2} &1 &4 &1 &2 &4 &0    \\ 
\hline
1  &30&11 &31 &11 &0 &0 &6 &15 &10 &11 &12 & \textbf{1}   \\
2  &35 &11 &36 &15 &0 &0 &7 &19 &11 &8 &11 &0    \\
3  &43 &16 &54 &21 &0 &0 &10 &17 &16 &14 &16 &0    \\
4  &53 &18 &63 &19 &0 &0 &1 &25 &14 &16 &16 &0    \\
5  &\textbf{71} &28 &\textbf{77} &\textbf{37} &0 &0 &\textbf{25} &\textbf{46} &24 &20 &\textbf{28} &0    \\
6  &56 &\textbf{30} &68 &27 &0 & 0&0 &38 &\textbf{27} &\textbf{23} &25 &0    \\
\hline
all  &27 &11 & 27&10 &0 &0 &0 &14 &11 &11 &0 &0    \\
\hline
\end{tabular}
\label{tab:aft}
\end{table}
\begin{figure}[t!]
\centering
\includegraphics[width=0.48\textwidth]{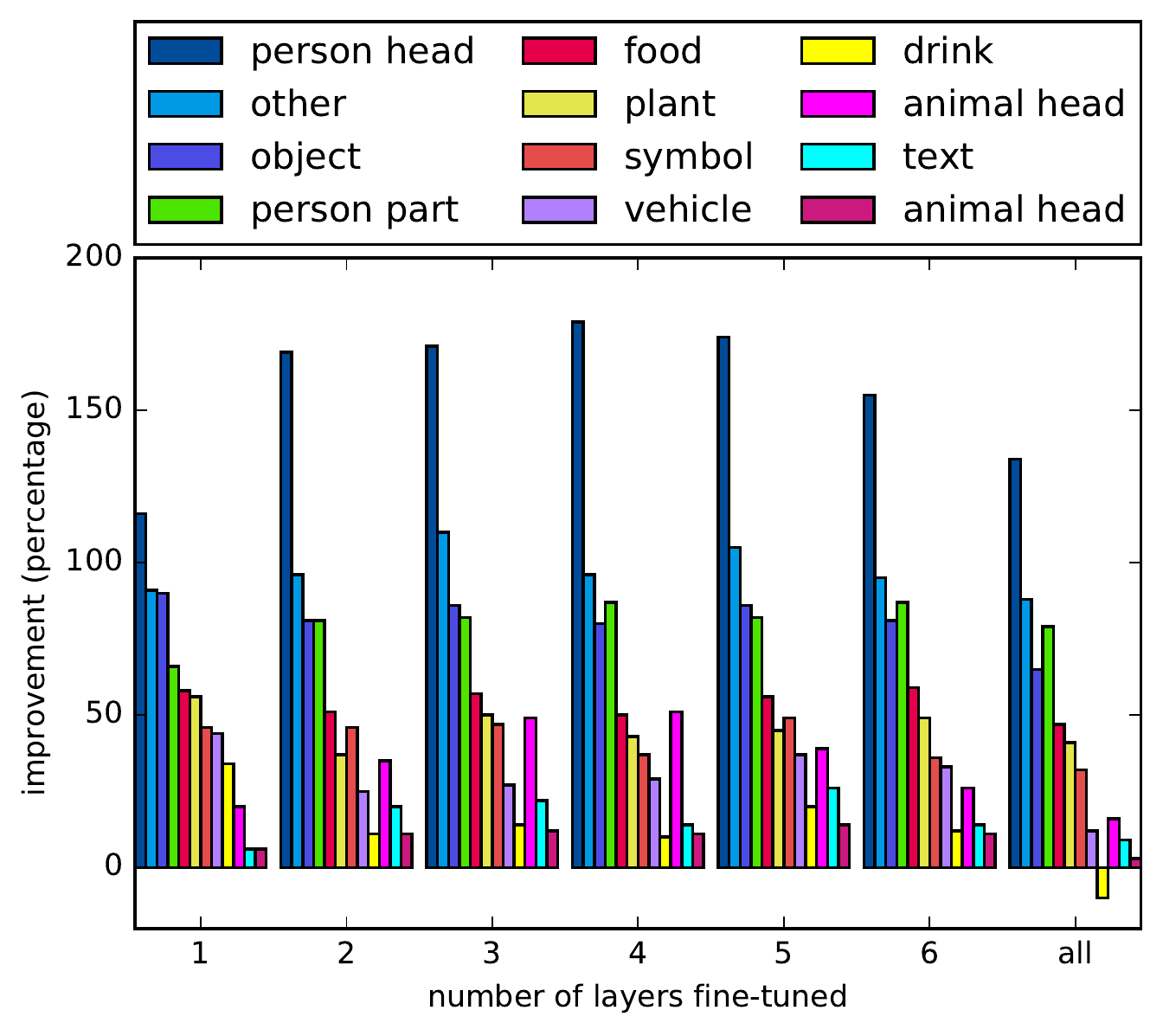}
\vspace{-17pt}
\caption{Top 10 mean NSS improvements for each category when fine-tuning different numbers of layers}
\label{fig:impro}
\end{figure}

\begin{figure}[t!]
\centering
\begin{subfigure}[b]{0.32\linewidth}
\includegraphics[width=\linewidth]{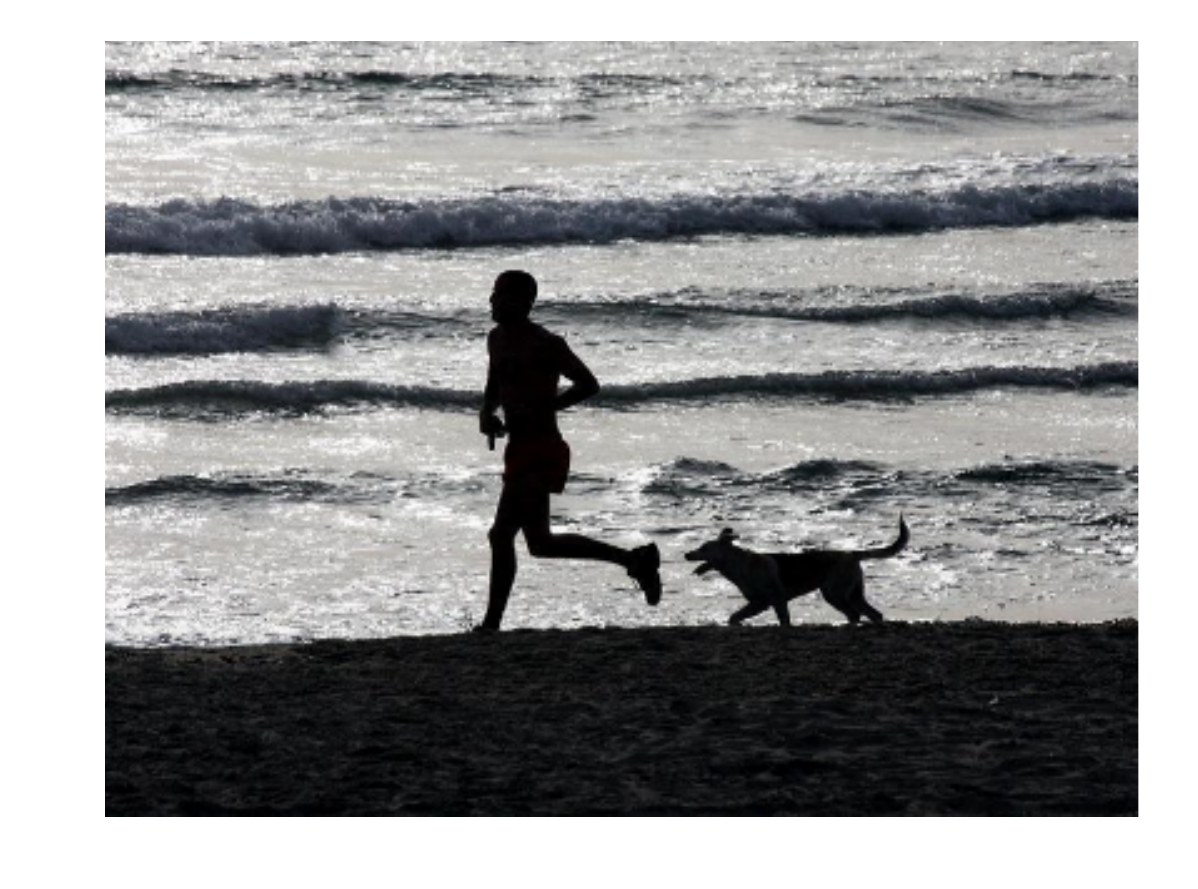}
\end{subfigure}
\begin{subfigure}[b]{0.32\linewidth}
\includegraphics[width=\linewidth]{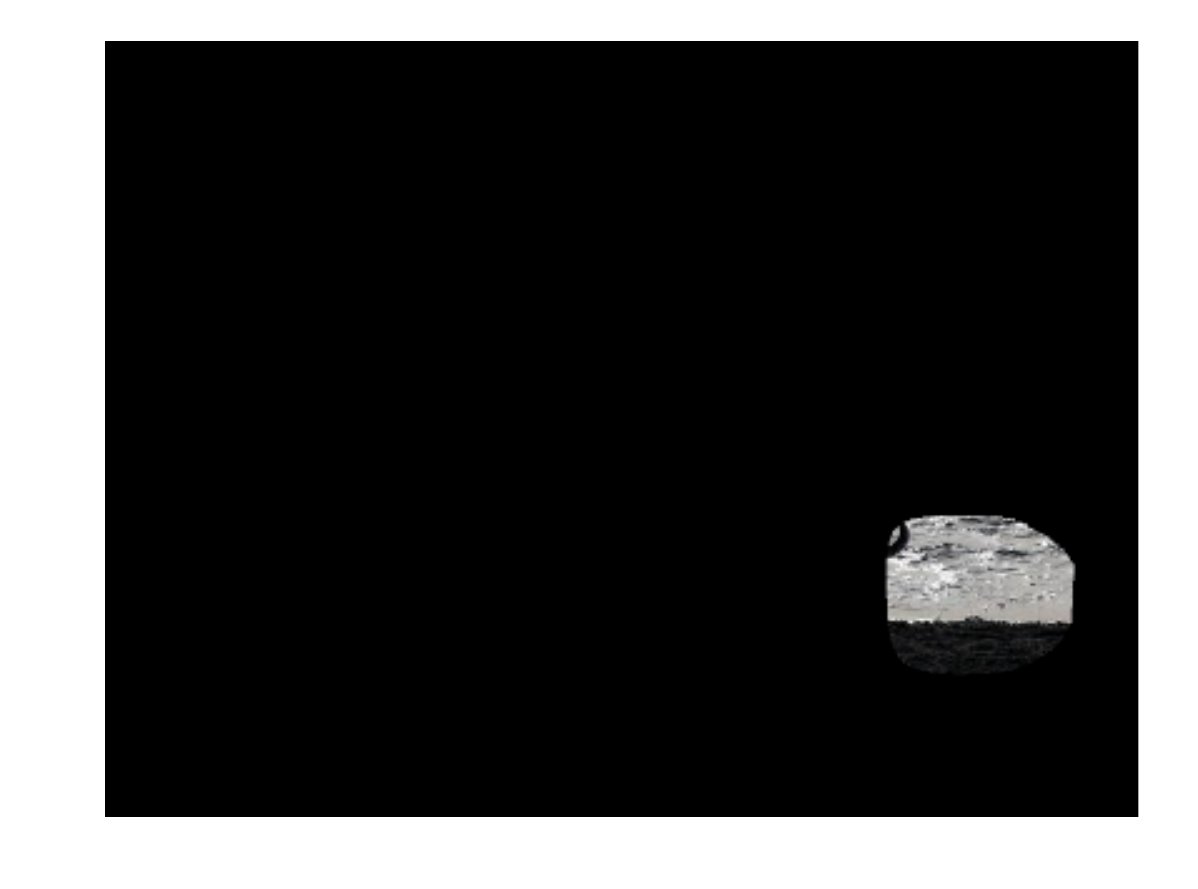}
\end{subfigure}
\begin{subfigure}[b]{0.32\linewidth}
\includegraphics[width=\linewidth]{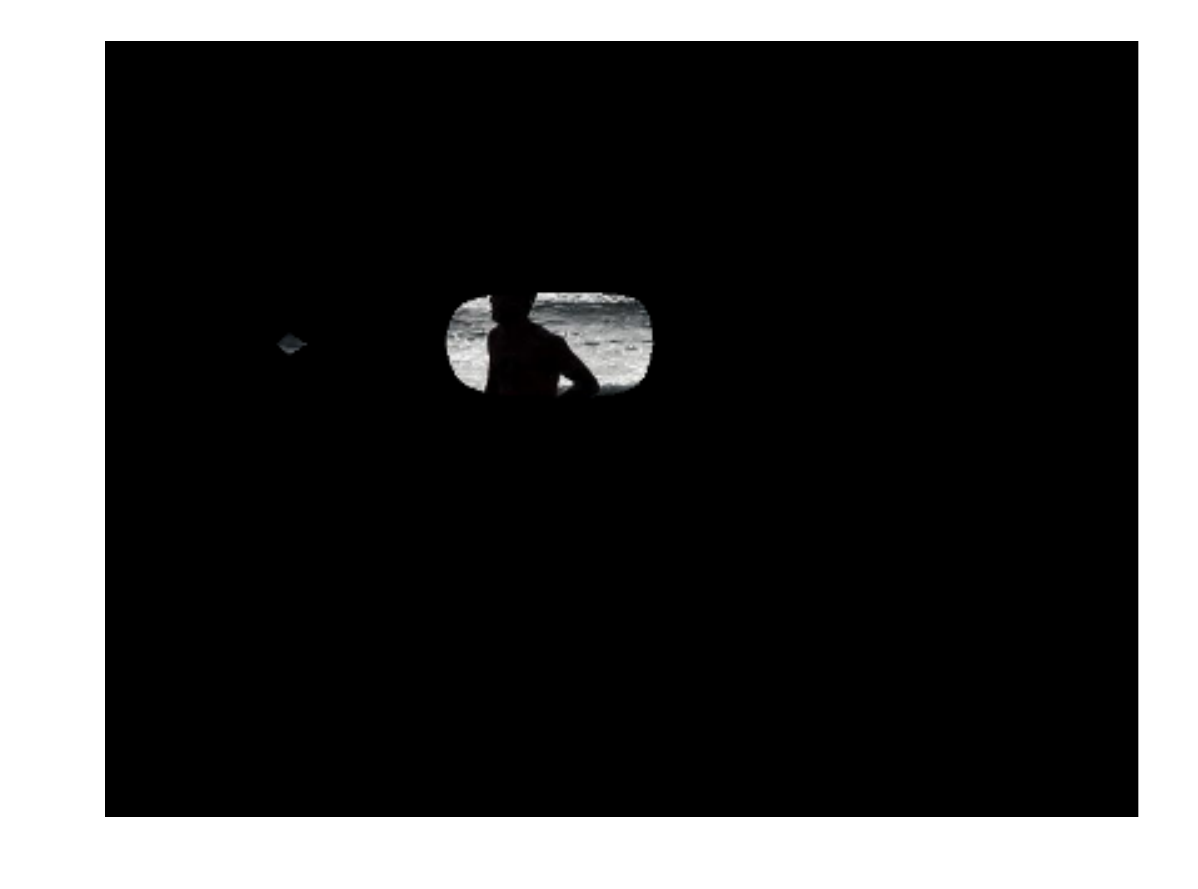}
\end{subfigure}
\begin{subfigure}[b]{0.32\linewidth}
\includegraphics[width=\linewidth]{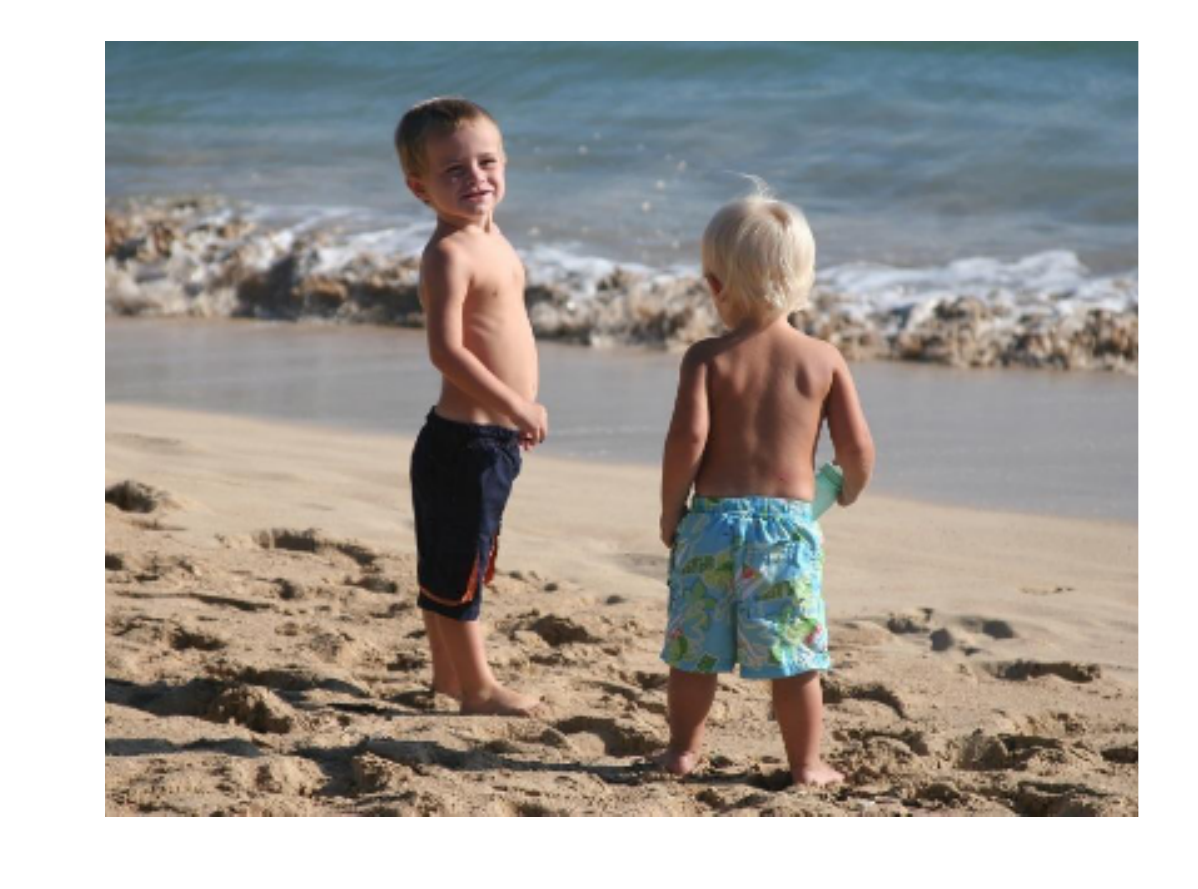}
\end{subfigure}
\begin{subfigure}[b]{0.32\linewidth}
\includegraphics[width=\linewidth]{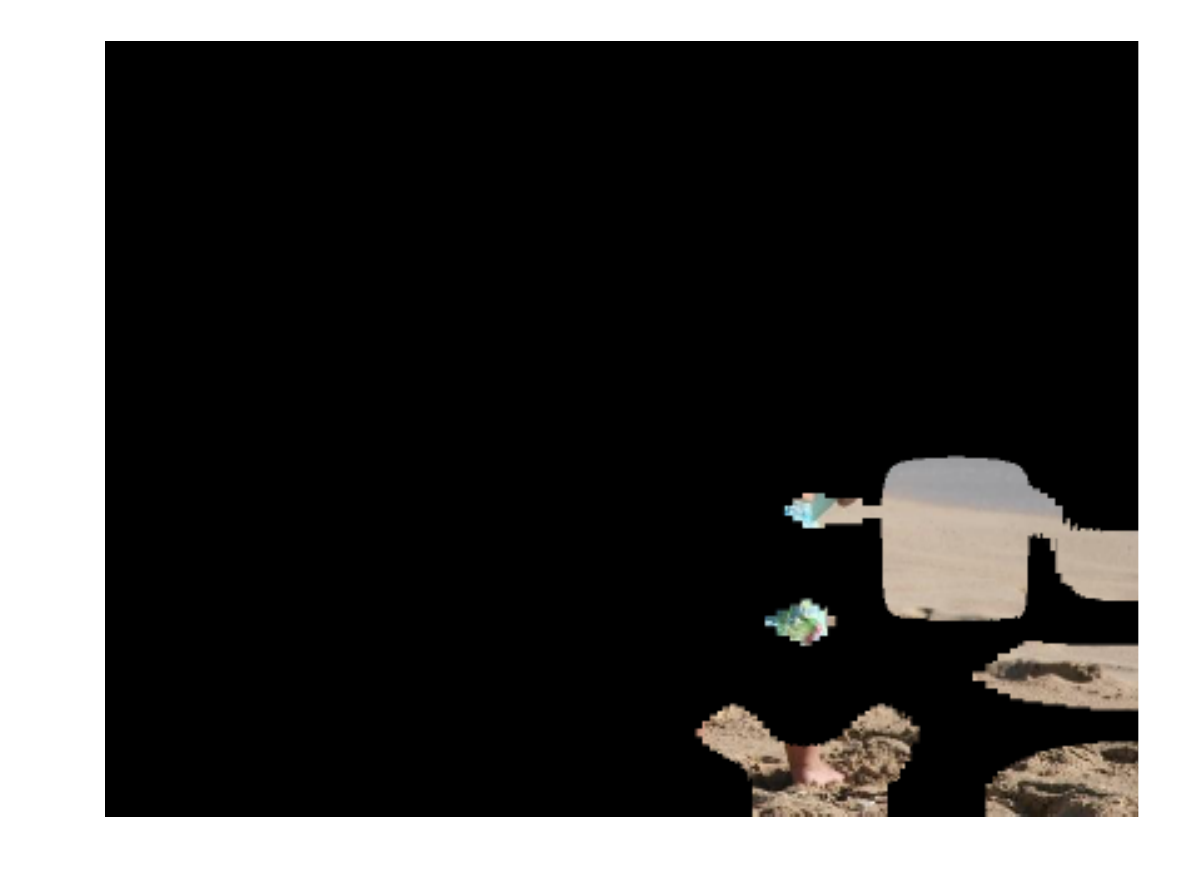}
\end{subfigure}
\begin{subfigure}[b]{0.32\linewidth}
\includegraphics[width=\linewidth]{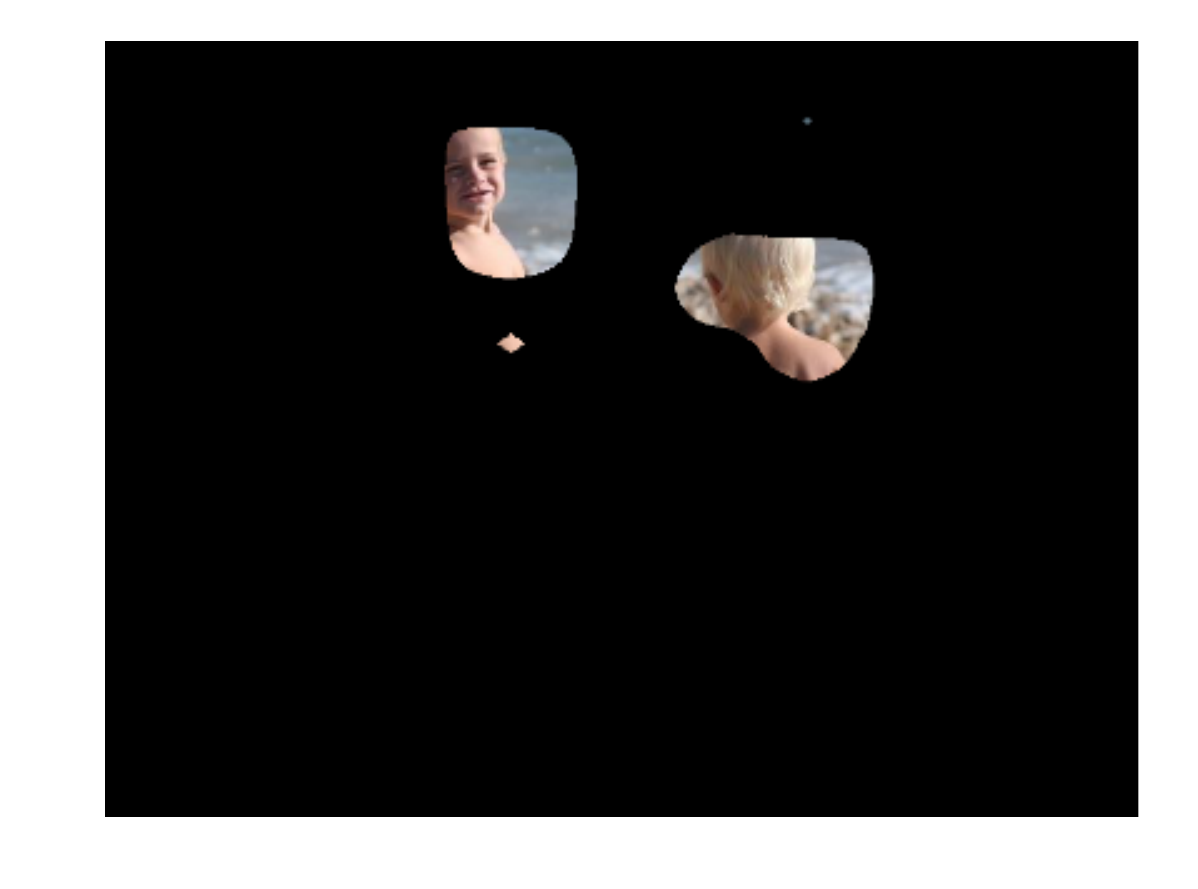}
\end{subfigure}
\caption{The $288th$ activation map in layer conv5-3 before and after fine-tuning for an input image. Notice that it becomes attuned to \textit{person head} after fine-tuning.}
\label{fig:aft}
\end{figure}

\begin{figure}[t!]
\centering
\begin{subfigure}[b]{0.24\linewidth}
\includegraphics[width=\linewidth]{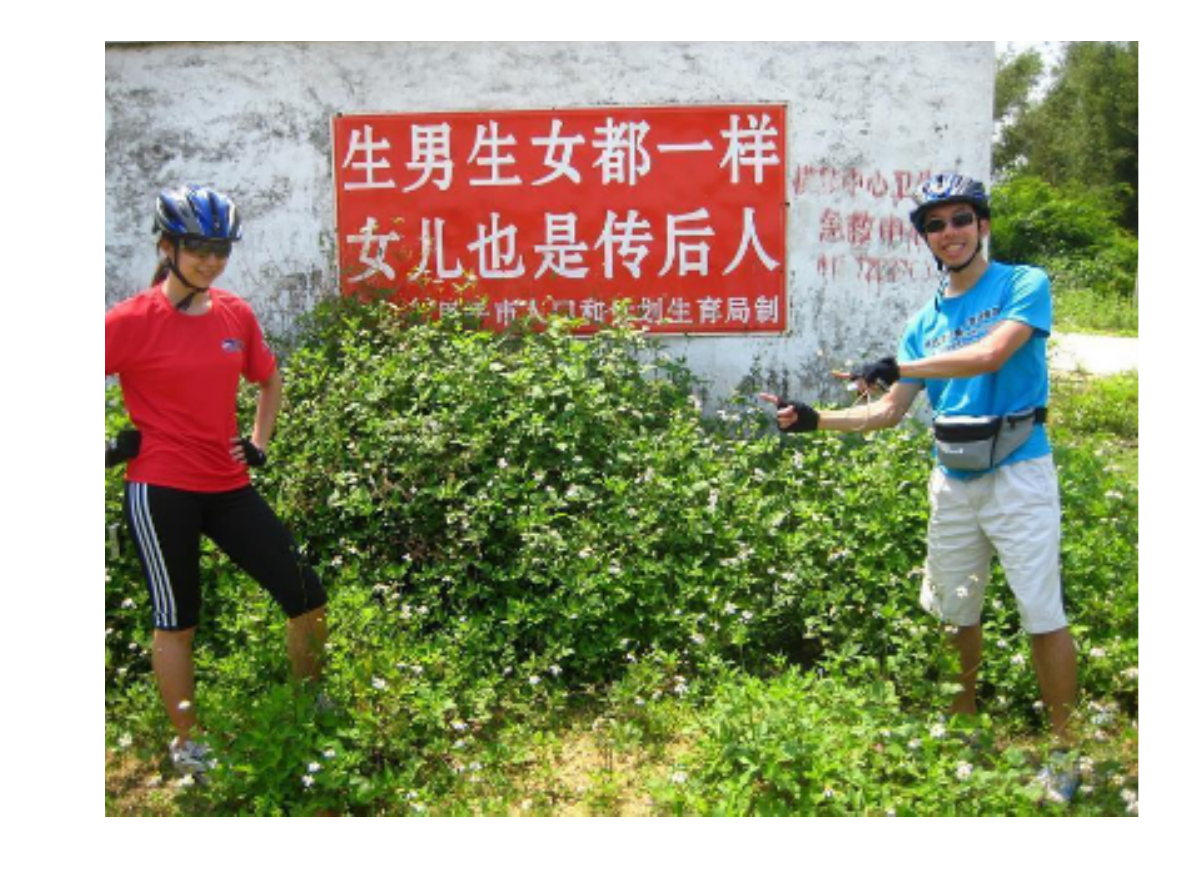}
\end{subfigure}
\begin{subfigure}[b]{0.24\linewidth}
\includegraphics[width=\linewidth]{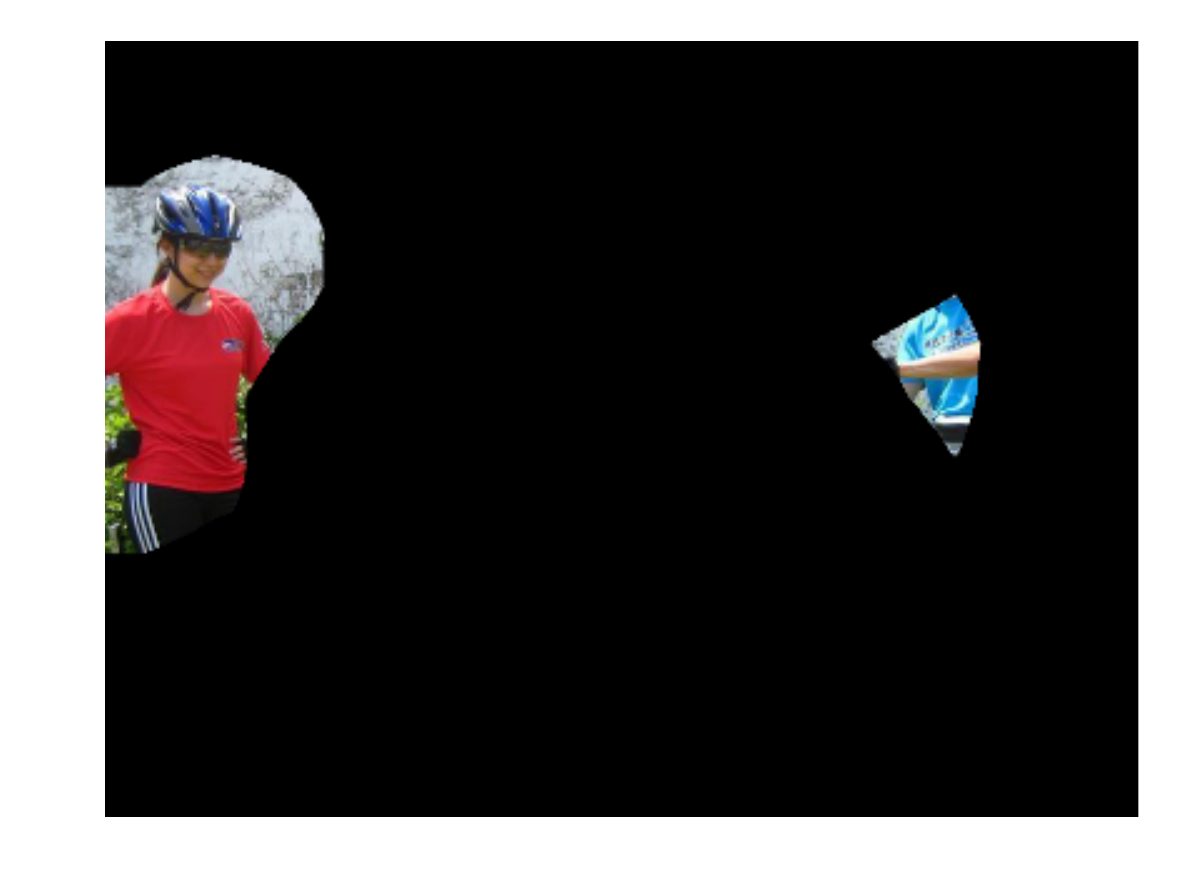}
\end{subfigure}
\begin{subfigure}[b]{0.24\linewidth}
\includegraphics[width=\linewidth]{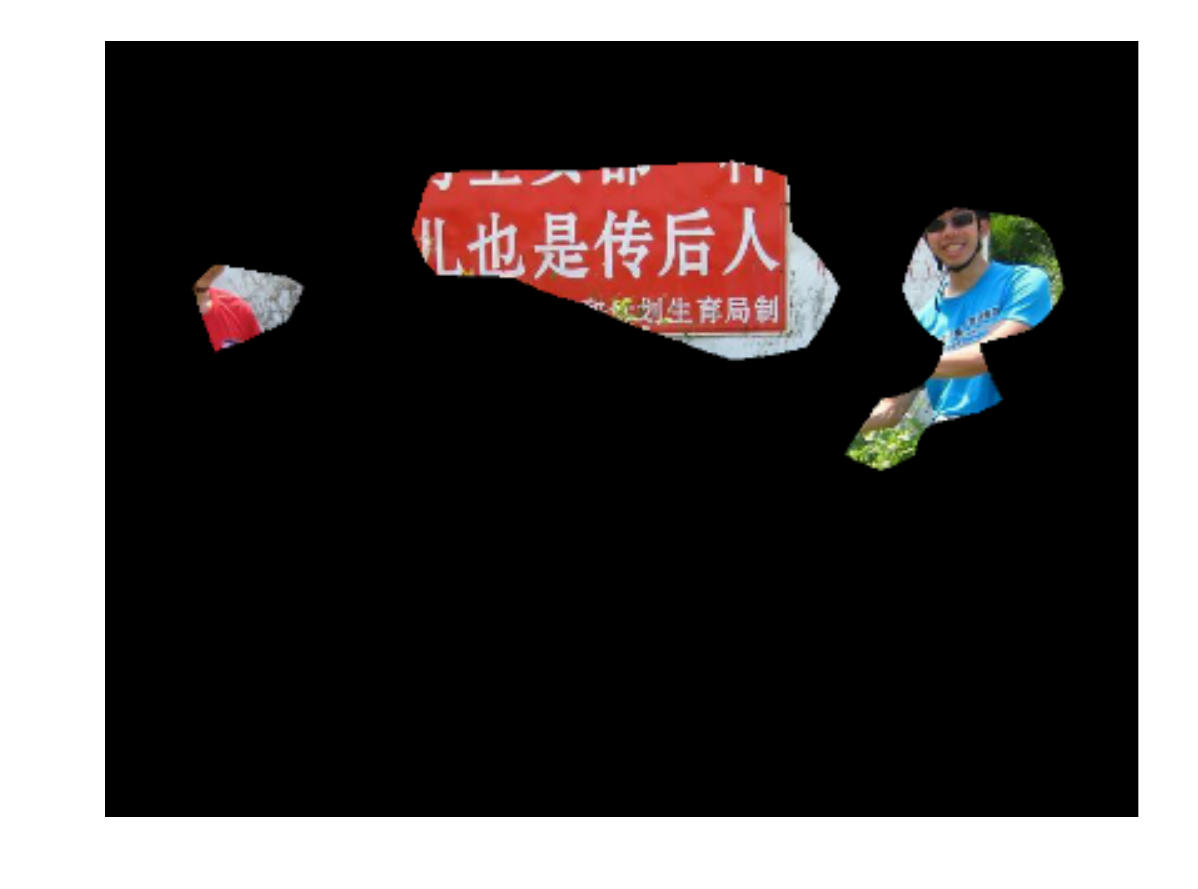}
\end{subfigure}
\begin{subfigure}[b]{0.24\linewidth}
\includegraphics[width=\linewidth]{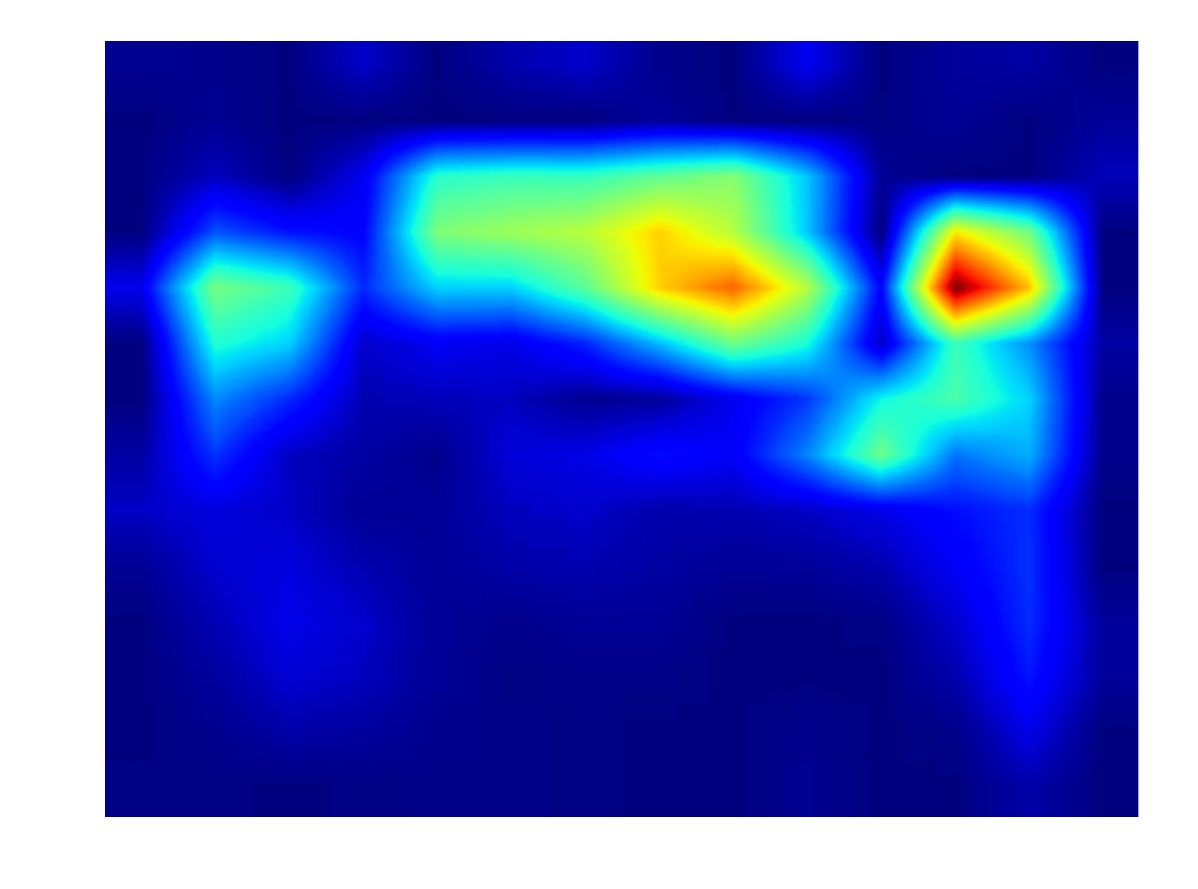}
\end{subfigure}
\begin{subfigure}[b]{0.24\linewidth}
\includegraphics[width=\linewidth]{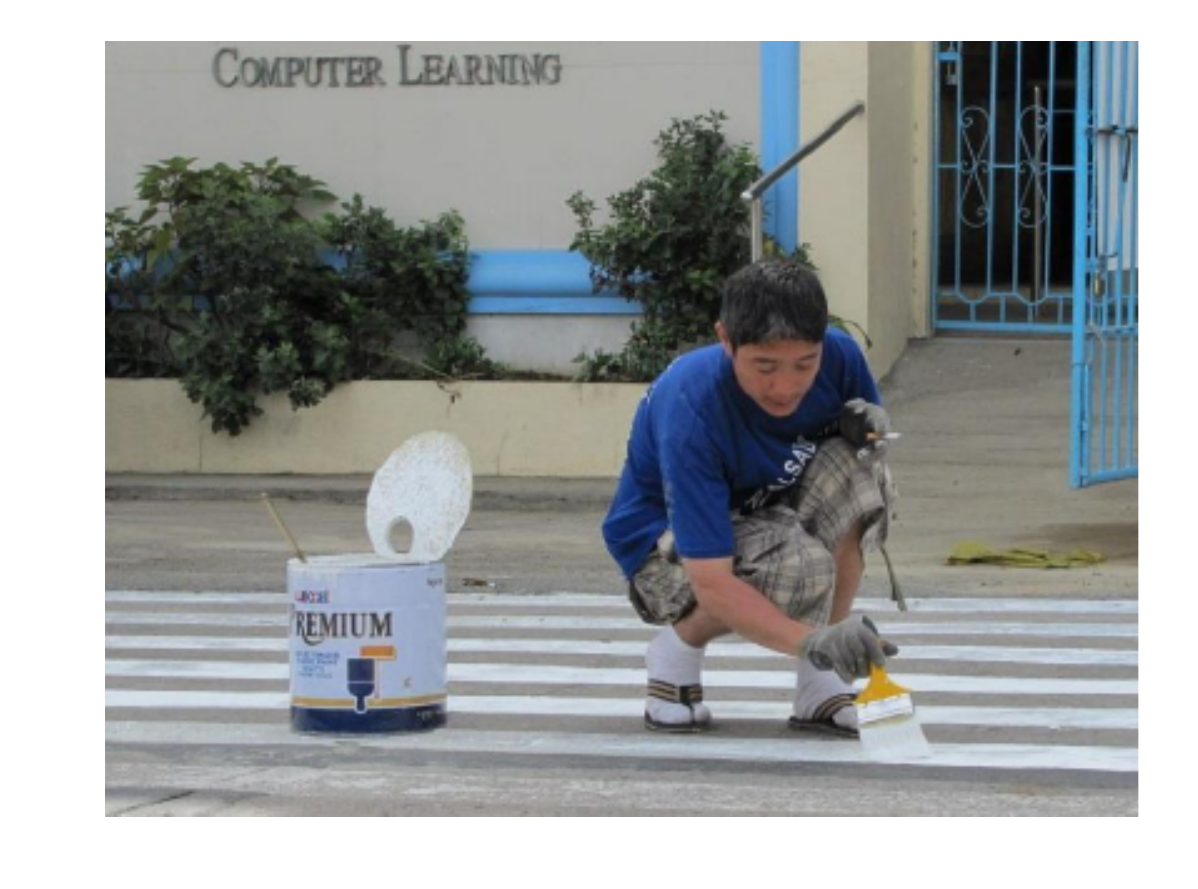}
\caption{}
\end{subfigure}
\begin{subfigure}[b]{0.24\linewidth}
\includegraphics[width=\linewidth]{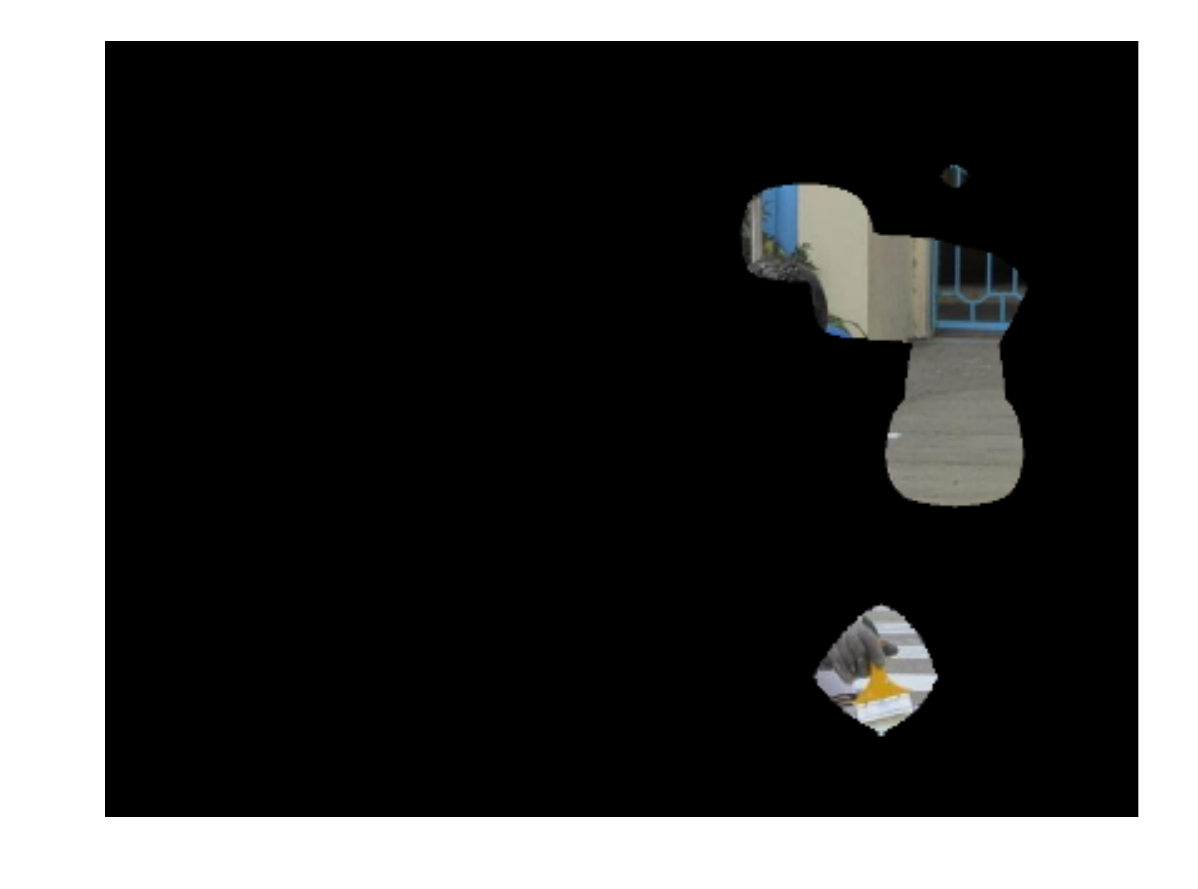}
\caption{}
\end{subfigure}
\begin{subfigure}[b]{0.24\linewidth}
\includegraphics[width=\linewidth]{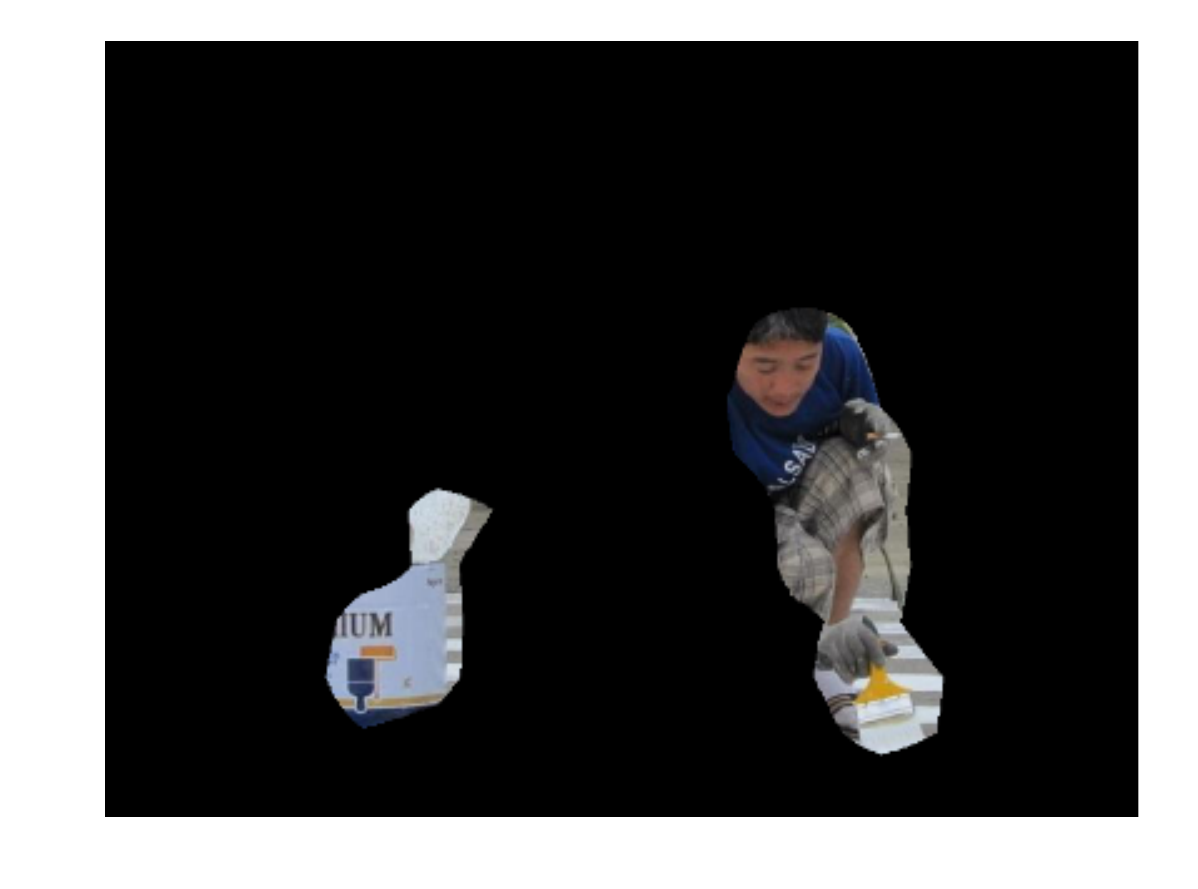}
\caption{}
\end{subfigure}
\begin{subfigure}[b]{0.24\linewidth}
\includegraphics[width=\linewidth]{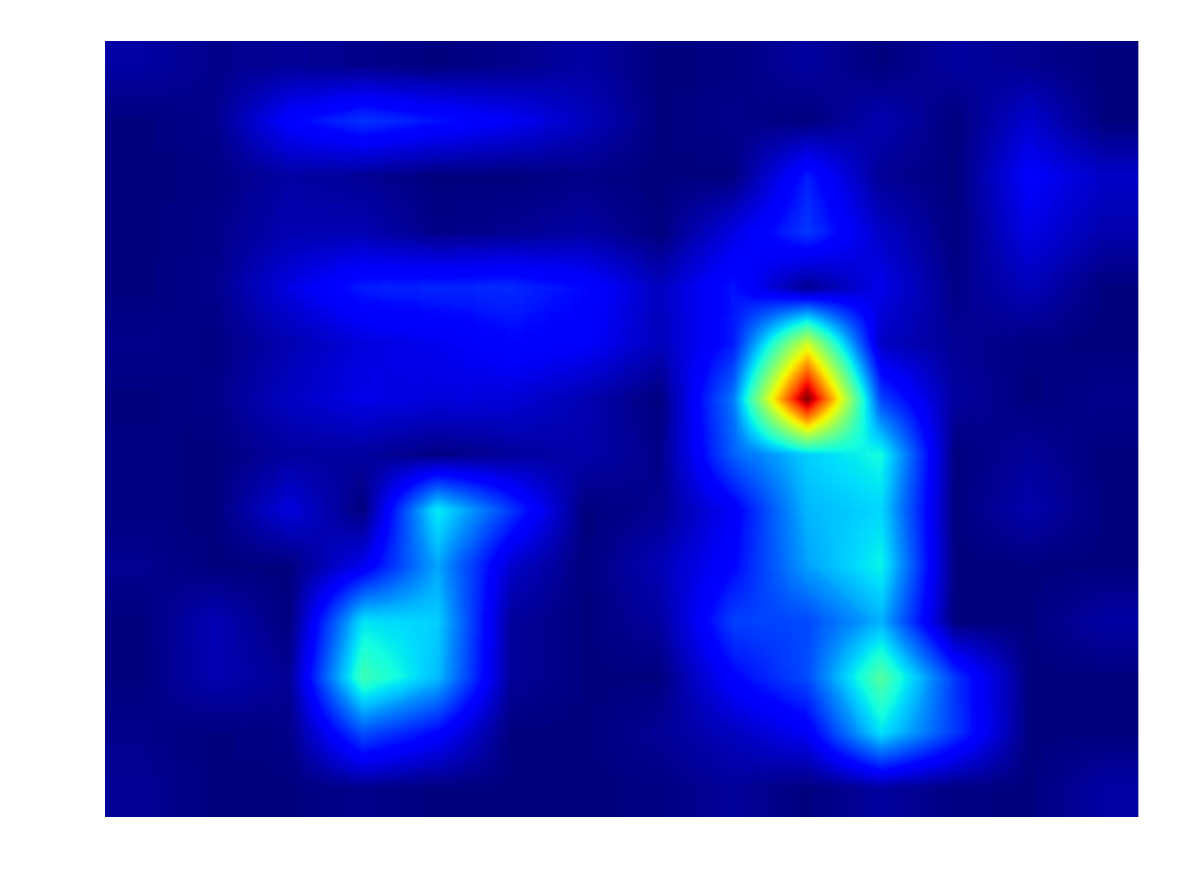}
\caption{}
\end{subfigure}
\caption{The $512th$ activation map in layer conv5-3 before and after fine-tuning (tuning 3 layers). (a) the input image, image overlapped with the activation map before fine-tuning (b), after fine-tuning (c), and the activation map after fine-tuning (d). After fine-tuning, despite the fact that this activation map has the highest mean NSS score for all regions annotated as \textit{text} in the dataset, it still favors \textit{heads}.}
\label{fig:t_aft}
\vspace{-15pt}
\end{figure}

\subsection{Saliency representation after fine-tuning}

Table~\ref{tab:aft} reports the statistics of visual saliency in layer conv5-3, which is directly used for saliency prediction, after fine-tuning different number of layers in the model (0 indicates that only the final $1 \times 1$ convolutional layer above the pre-trained model for regression was trained on saliency data, without fine-tuning the pre-trained model; 1 means fine-tuning the layer conv5-3, 2 means fine-tuning the layers conv5-3 and conv5-2, and so on).

From Table~\ref{tab:aft}, we can see that after fine-tuning, the activation maps became more selective to visual saliency, as the mean NSS values improve for all saliency categories. The improvements are, however, uneven as can be seen in Fig.~\ref{fig:impro}; \ie some categories improve more. For example, \textit{person head} improves the most from 1.21 to 3.38 when fine-tuning 4 layers (see Fig.~\ref{fig:aft}), while for \textit{text}, the improvement is relatively small (at most from 1.1 to 1.38 when fine-tuning 5 layers). 
Similar effects can be seen in Fig.~\ref{fig:t_aft}, where activation maps which are most selective to texture regions in the image after fine-tuning still respond to \textit{head} regions in the image, and with high activation values. It is withstanding that an ANOVA test and multiple comparisons indicate person head, animal head, and other are significantly different than all other categories and each other.

An interesting observation is that by fine-tuning more layers, \ie more than 3 layers (and especially when fine-tuning all the layers), the mean NSS values or the number of activation maps for some saliency categories start to decrease. We also do not gain more saliency prediction improvement by fine-tuning more layers. We speculate that one reason behind this observation might be the quality and quantity of the current saliency data, which is biased towards specific salient objects and regions.

\begin{figure}[t!]
\centering
\begin{subfigure}[b]{0.48\linewidth}
\includegraphics[width=\linewidth]{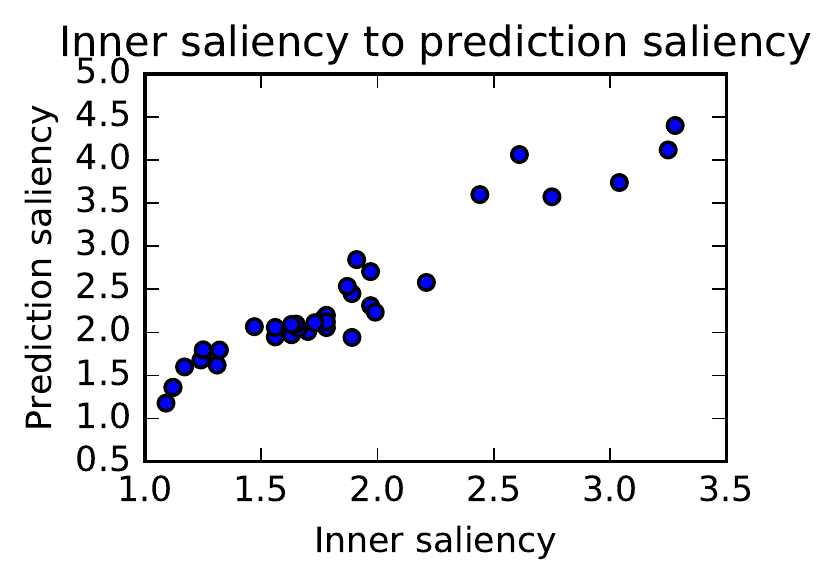}
\vspace{.5pt}
\end{subfigure}
\begin{subfigure}[b]{0.48\linewidth}
\includegraphics[width=\linewidth]{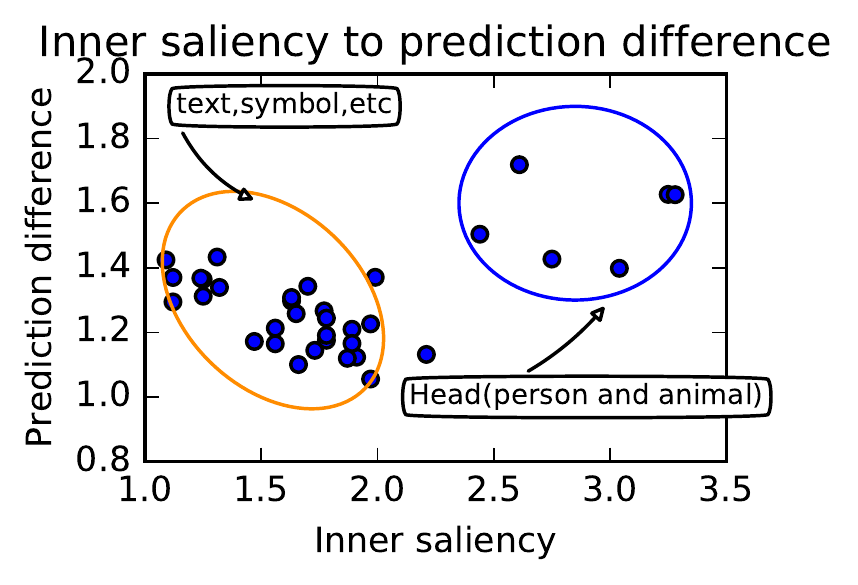}
 \vspace{.5pt}
\end{subfigure}

\begin{subfigure}[b]{0.32\linewidth}
\includegraphics[width=\linewidth]{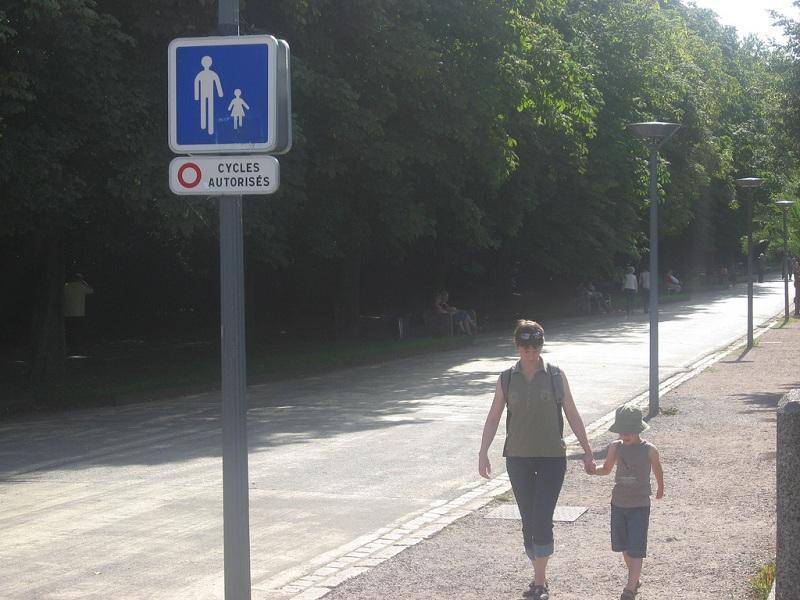}
\end{subfigure}
\begin{subfigure}[b]{0.32\linewidth}
\includegraphics[width=\linewidth]{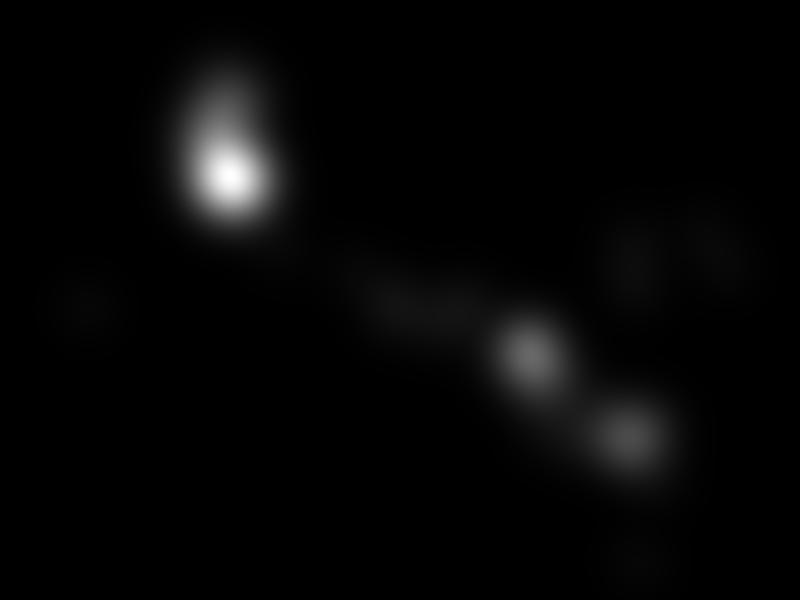}
\end{subfigure}
\begin{subfigure}[b]{0.32\linewidth}
\includegraphics[width=\linewidth]{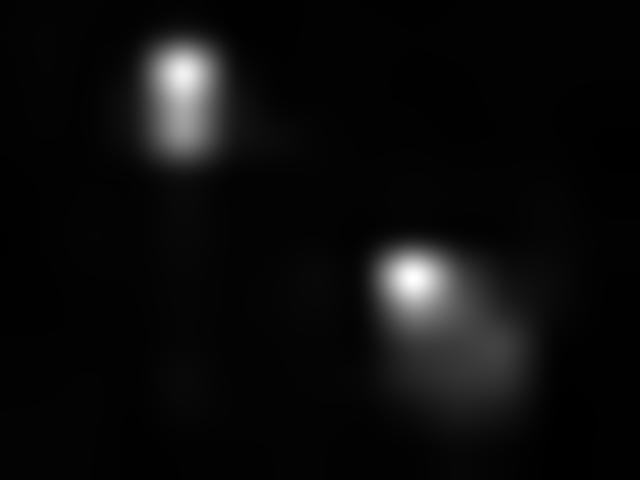}
\end{subfigure}
\caption{Top row: the relationship between the inner representation saliency and output saliency (left) and the output prediction difference (right). Bottom row: an example image, ground truth fixation map, and model prediction.}
\label{fig:rela}
\end{figure}

\section{The Relationship between Intermediate and Final Representations}
What is the relation between a model's inner representation and its output? In other words, are the categories that are more salient in the inner representations also more active in the output saliency map? To what extend the salient categories within the inner representations agree with the ground truth salient categories? To answer these questions, we define the the inner saliency for each category as the mean NSS value of the top 10 activation maps for it, which is the same as the values in Table~\ref{tab:aft}. Following steps of~\cite{Tavakoli_2017_CVPR} for fine-grained contextual analysis,
we assign the output saliency of a model, defined as $OS_c$ for category $c$, as the output's mean NSS score of all salient regions in the OSIE-SR dataset that belong to category $c$,
\begin{align}
OS_c = \frac{1}{A} \sum_{i=1}^{N} \sum_{j=1}^{M} NSS( pred_i, mask_{i,c_j}\boldsymbol{\cdot} fix_i)
\end{align}

\noindent where, $A$ is the total number of salient regions for category $c$, $N$ is the total number of images, and $M$ is total number of regions for category $c$ in $i_{th}$ image, $pred_i$ is the model prediction for  $i_{th}$ image, $mask_{i,c_j}$ is the annotated mask for the $j_{th}$ region in $i_{th}$ image for category $c$ (if category $c$ exists in the image) and $fix_i$ is the fixation location on the $i_{th}$ image.

To measure the relation between salient categories and representations, we define the concept of output salient category difference, denoted as $OD_c$. It measures the mean difference of the NSS score between a model prediction and the ground truth with respect to the salient categories,
\begin{align}
OD_c = \frac{1}{A} \sum_{i=1}^{A} \sum_{j=1}^{M} \abs{f_{c_j}(pred_i)-f_{c_j}(GT_i)}
\end{align}

\noindent where, $f_{c_j}(pred_i)=NSS( pred_i, mask_{i,c_j}\boldsymbol{\cdot} fix_i)$, and $GT_i$ is the ground truth saliency map.

The results are shown in Fig.~\ref{fig:rela}. As depicted in the top left of this figure, the model's saliency output is correlated to the saliency of inner representations. In other words, we can see that if a category is more salient in the model's inner representation, it is also more salient in the model's output (Spearman's correlation coefficient: $r_s=0.96$). 
The salient categories in inner representations and output, however, migh be different. The output salience category difference is lower for less salient inner categories and higher for more salient inner ones. For example, as depicted in top right panel of Fig.~\ref{fig:rela}, the \textit{head} category has more salient inner representations and higher output salient category difference (ANOVA test with measurements and clusters as factors showing significant difference $p=9e^{-9}<0.05$). In other words, the model has learned to fire on faces, irrespective of whether they are salient in the context of the given image.

\section{Model Performance and Representations over Synthetic Search Arrays}
\begin{figure}[h!]
\centering
\begin{subfigure}[b]{0.24\linewidth}
\includegraphics[width=\linewidth]{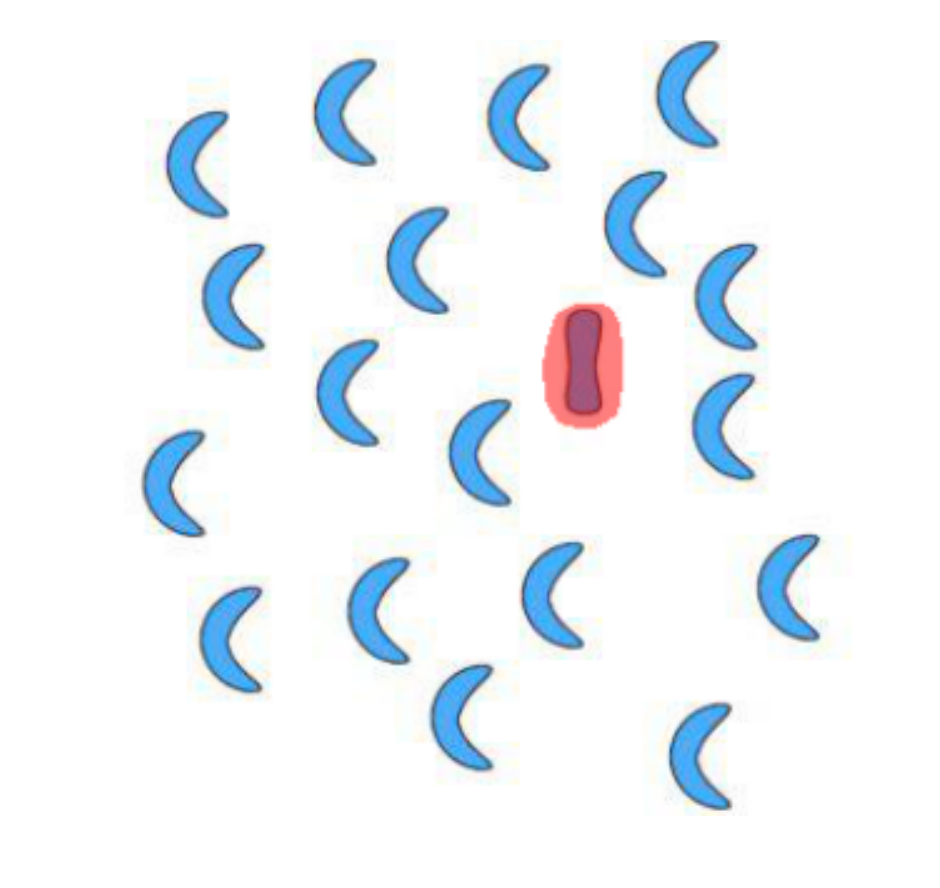}
\end{subfigure}
\begin{subfigure}[b]{0.24\linewidth}
\includegraphics[width=\linewidth]{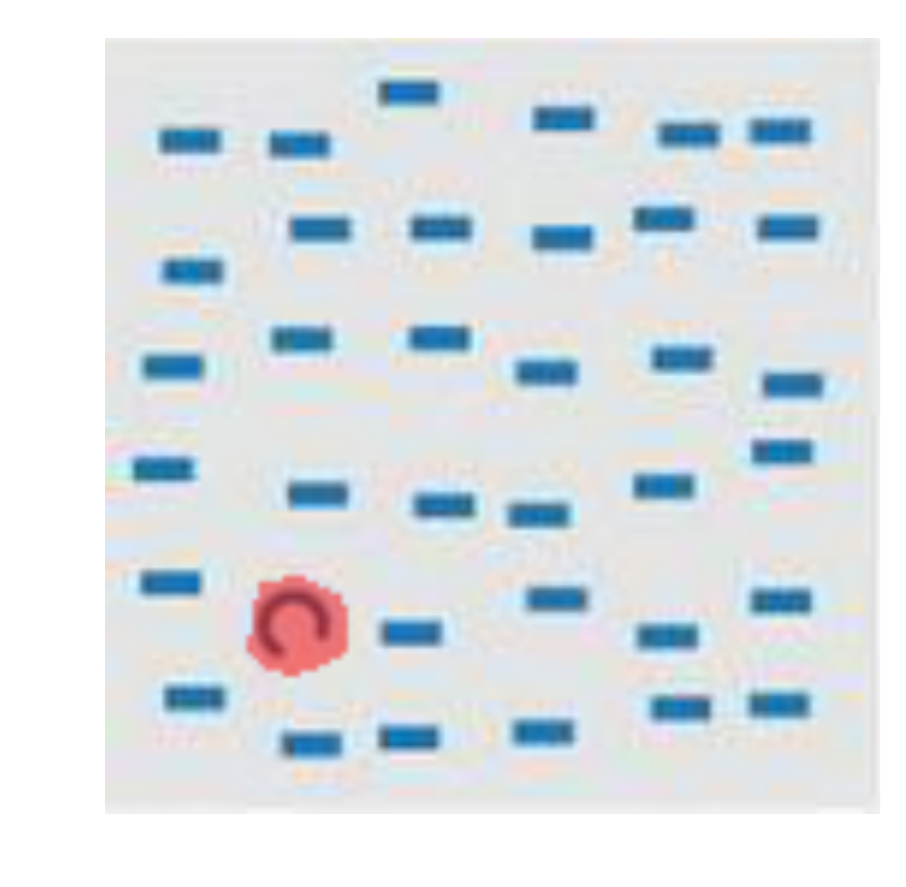}
\end{subfigure}
\begin{subfigure}[b]{0.24\linewidth}
\includegraphics[width=\linewidth]{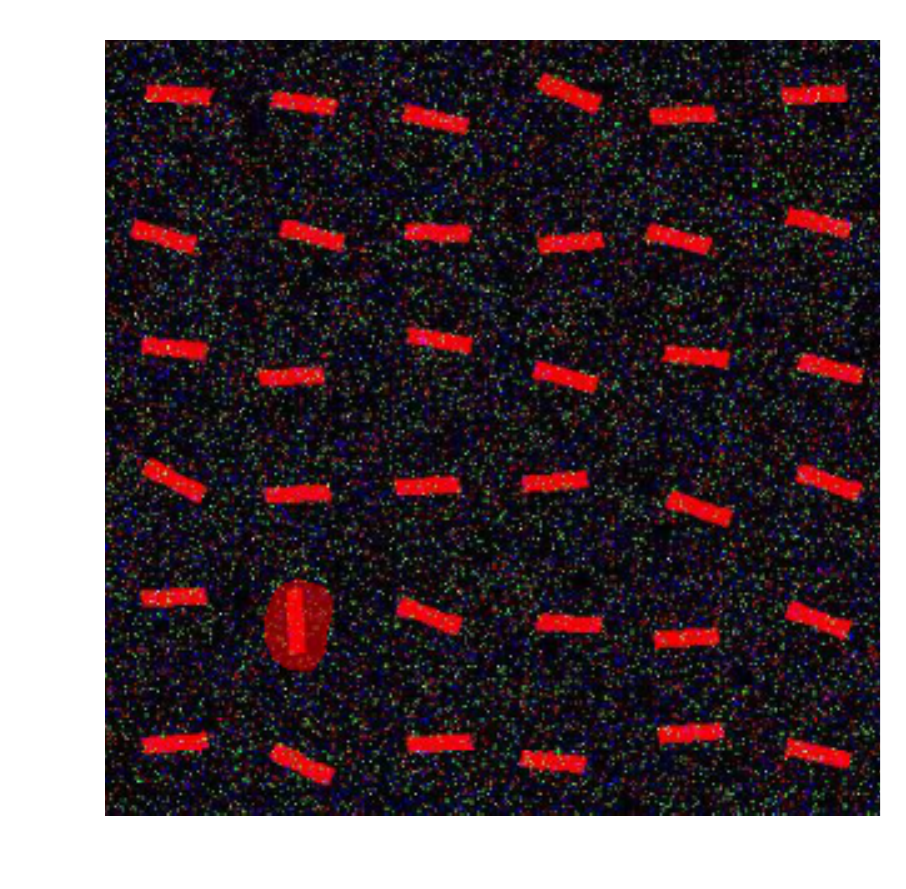}
\end{subfigure}
\begin{subfigure}[b]{0.24\linewidth}
\includegraphics[width=\linewidth]{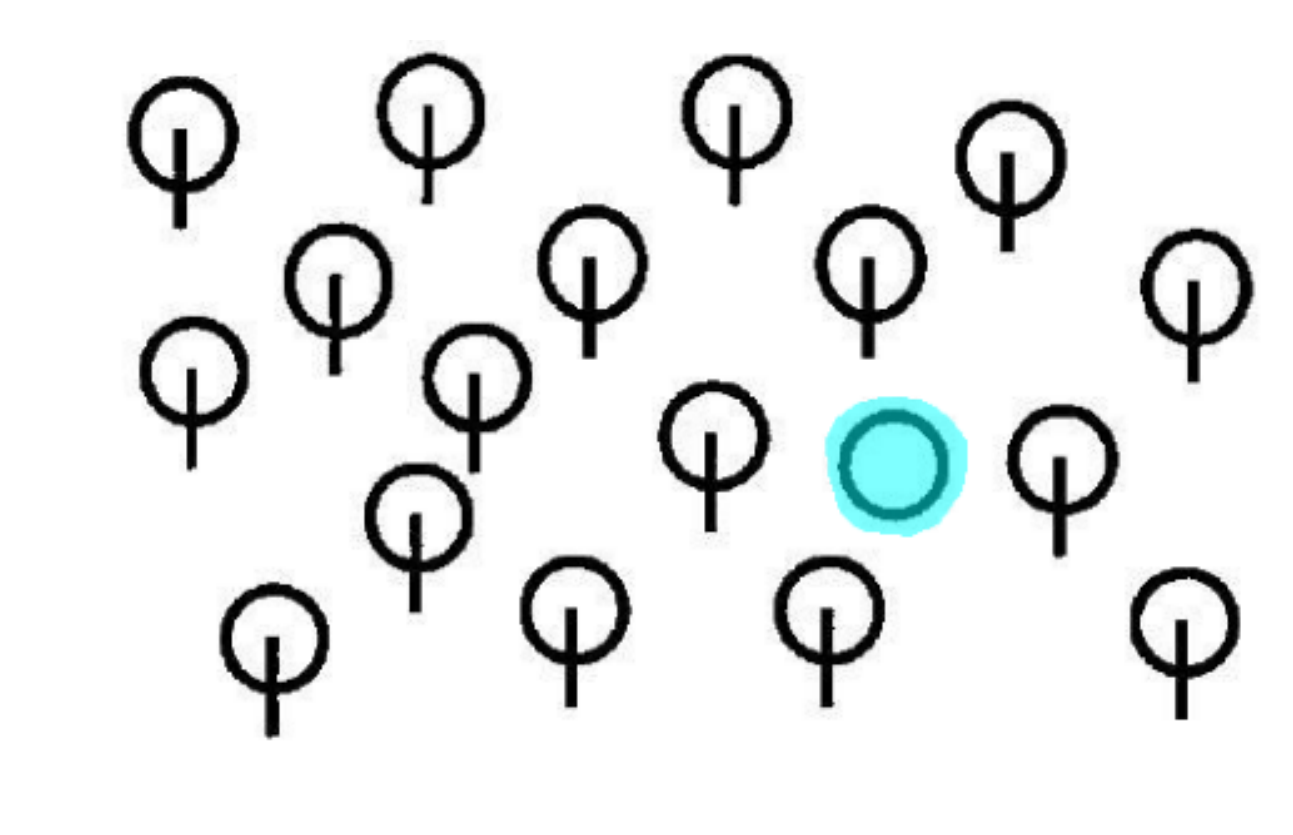}
\end{subfigure}
\begin{subfigure}[b]{0.24\linewidth}
\includegraphics[width=\linewidth]{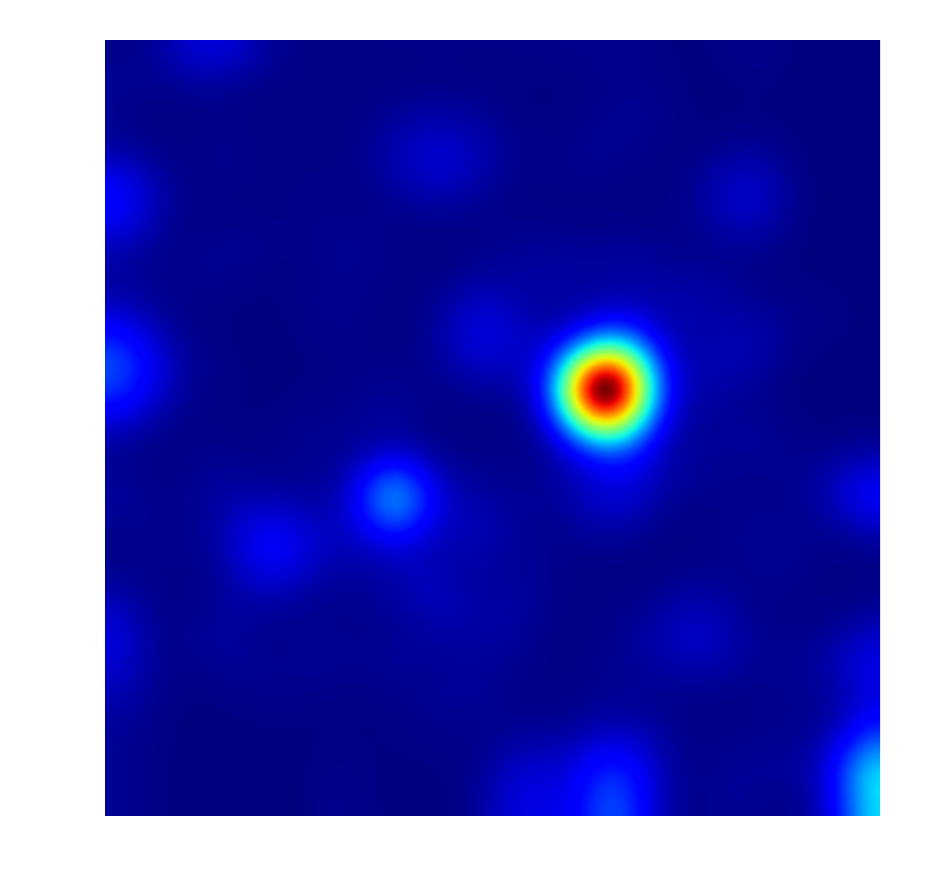}
\end{subfigure}
\begin{subfigure}[b]{0.24\linewidth}
\includegraphics[width=\linewidth]{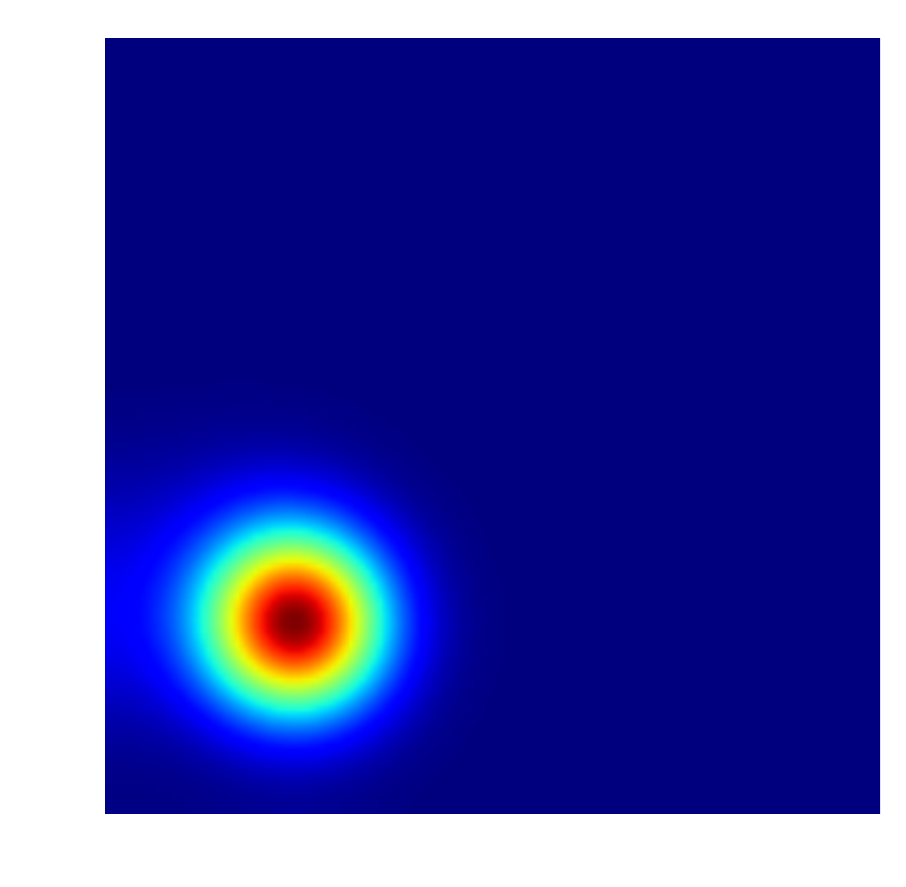}
\end{subfigure}
\begin{subfigure}[b]{0.24\linewidth}
\includegraphics[width=\linewidth]{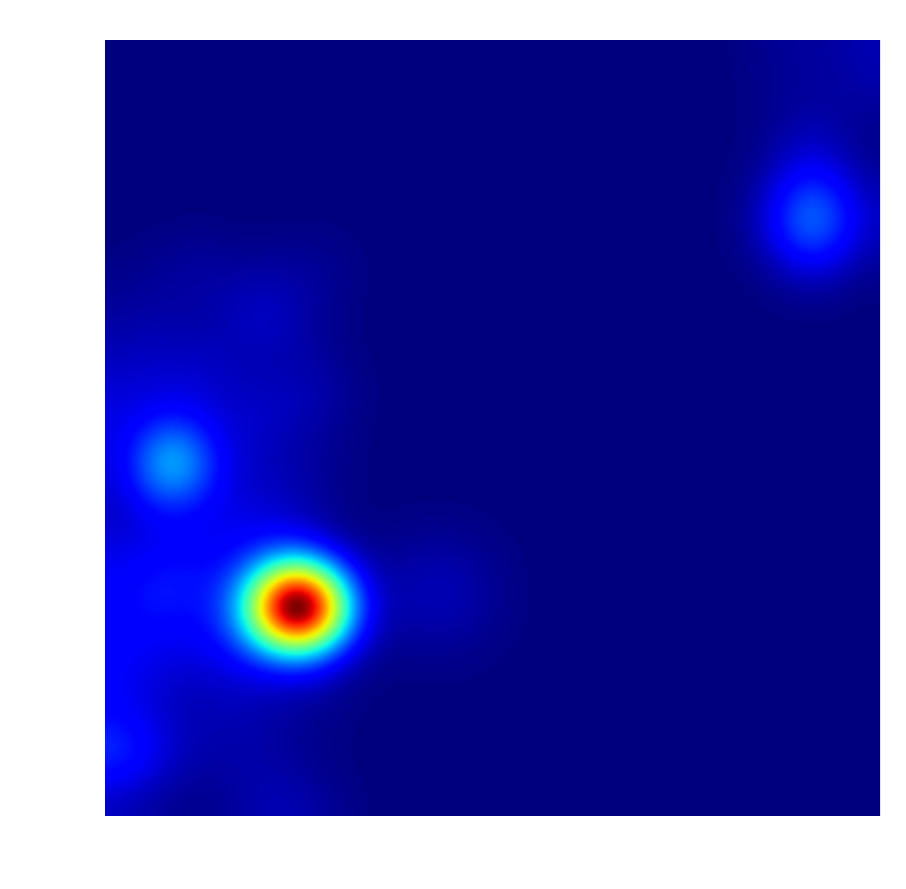}
\end{subfigure}
\begin{subfigure}[b]{0.24\linewidth}
\includegraphics[width=\linewidth]{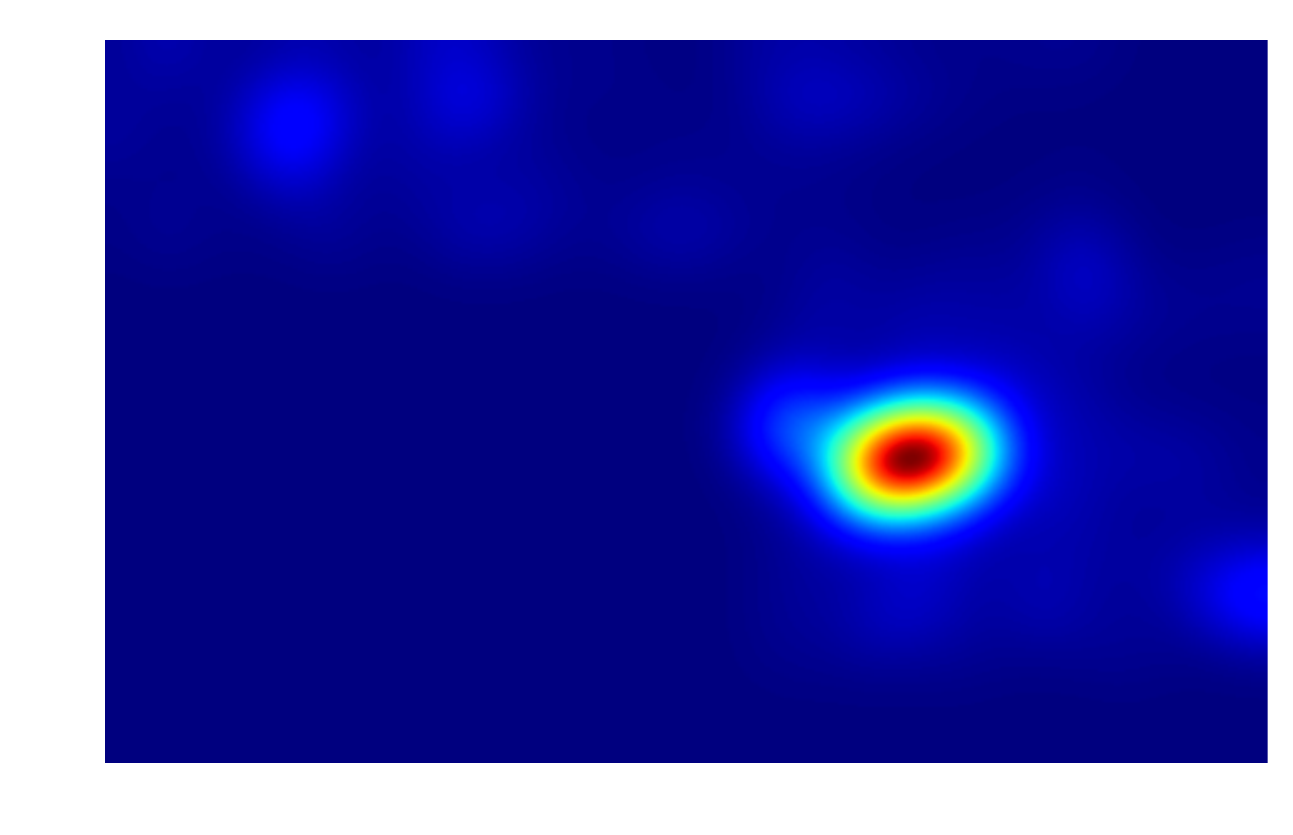}
\end{subfigure}
\caption{Top row: synthetic search arrays with their annotated masks. Bottom row: the activation maps from layer conv4-1 that best correlated with the masked regions in the synthetic images.}
\label{fig:syn_actm}
\end{figure}

How do deep saliency models perform over the synthetic images?
We compare the performance of deep and classical saliency models on a set of synthetic images. These images are designed to simulate the feature pop-out, and have been extensively used to study human attention, but they have not been considered for evaluating deep saliency models. There exists no eye-fixation on such images, but is it easy to locate the target/salient item in the array (in fact we manually labeled the images). we assess the performance of the models using Normalized Mean value under the annotated Mask (NMM) for each image:

\begin{align}
NMM_i = mean_{mask_i}(\frac{pred_i-\mu(pred_i)}{\sigma(pred_i)})
\end{align}

\noindent where $mask_i$ and $pred_i$ are the annotated mask and model prediction for the $i_{th}$ image, respectively.
\begin{table}[h!]
\caption{Performance comparison between deep models and classical models on synthetic images in terms of NMM.}
\small
\setlength{\tabcolsep}{2.5pt} 
\renewcommand{\arraystretch}{1.1} 
\vspace{3pt}
\centering
\begin{tabular}{|c|c|c||c|c|}

\hline
\multicolumn{3}{|c ||}{Deep models}&\multicolumn{2}{c |}{Classical models}\\
\hline
\hline

DeepGaze~\Romannum{2}~\cite{kummerer2017understanding}&SAM~\cite{cornia2018predicting}&DVA~\cite{wang2018deep}&GBVS~\cite{harel2007graph}&BMS~\cite{zhang2013saliency}\\
\hline

1.66&1.25&1.19&2.57&3.65\\
\hline
\end{tabular}
\label{tab:dvsc}
\end{table}
\begin{figure}[h!]
\centering
\begin{subfigure}[b]{0.15\linewidth}
\includegraphics[width=\linewidth]{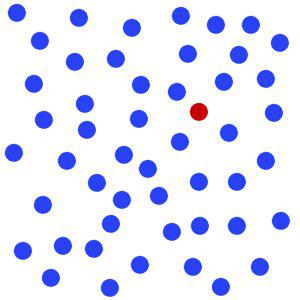}
\end{subfigure}
\begin{subfigure}[b]{0.15\linewidth}
\includegraphics[width=\linewidth]{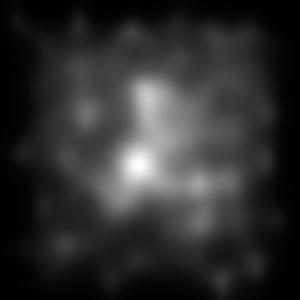}
\end{subfigure}
\begin{subfigure}[b]{0.15\linewidth}
\includegraphics[width=\linewidth]{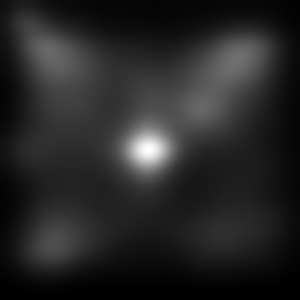}
\end{subfigure}
\begin{subfigure}[b]{0.15\linewidth}
\includegraphics[width=\linewidth]{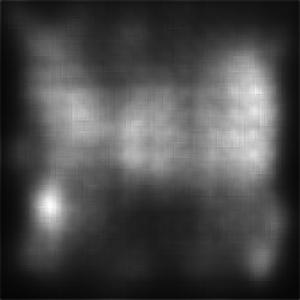}
\end{subfigure}
\begin{subfigure}[b]{0.15\linewidth}
\includegraphics[width=\linewidth]{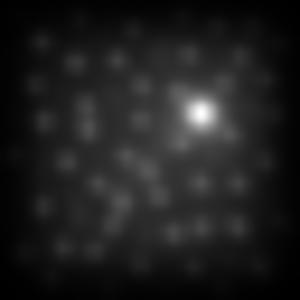}
\end{subfigure}
\begin{subfigure}[b]{0.15\linewidth}
\includegraphics[width=\linewidth]{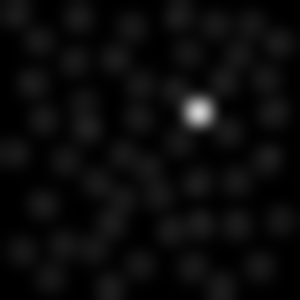}
\end{subfigure}
\begin{subfigure}[b]{0.15\linewidth}
\includegraphics[width=\linewidth]{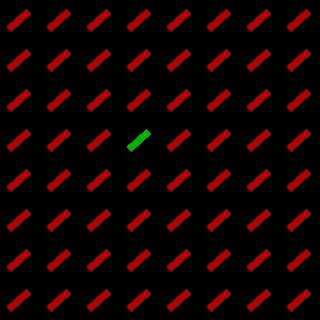}
\end{subfigure}
\begin{subfigure}[b]{0.15\linewidth}
\includegraphics[width=\linewidth]{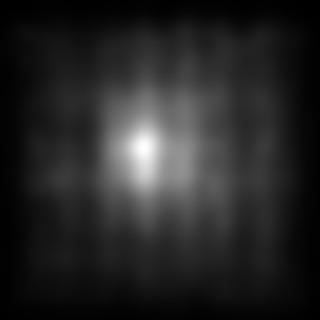}
\end{subfigure}
\begin{subfigure}[b]{0.15\linewidth}
\includegraphics[width=\linewidth]{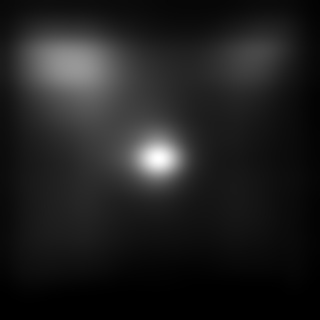}
\end{subfigure}
\begin{subfigure}[b]{0.15\linewidth}
\includegraphics[width=\linewidth]{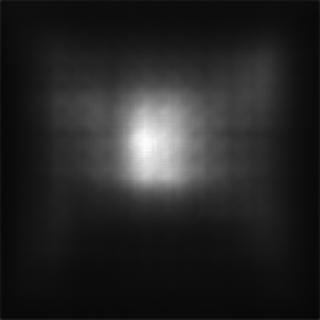}
\end{subfigure}
\begin{subfigure}[b]{0.15\linewidth}
\includegraphics[width=\linewidth]{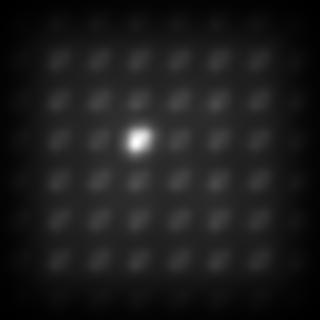}
\end{subfigure}
\begin{subfigure}[b]{0.15\linewidth}
\includegraphics[width=\linewidth]{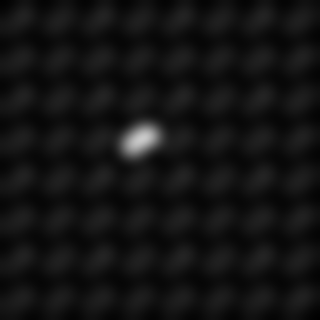}
\end{subfigure}
\begin{subfigure}[b]{0.15\linewidth}
\includegraphics[width=\linewidth]{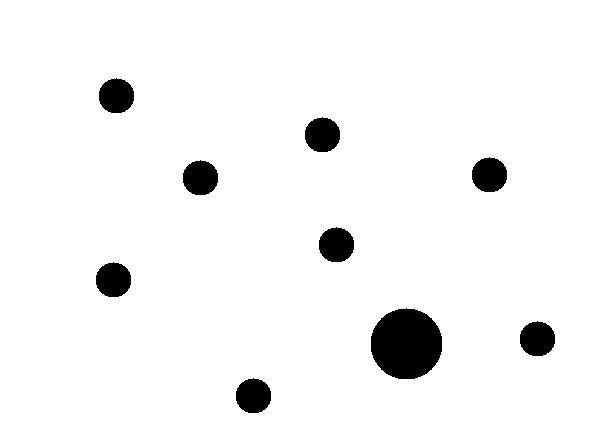}
\end{subfigure}
\begin{subfigure}[b]{0.15\linewidth}
\includegraphics[width=\linewidth]{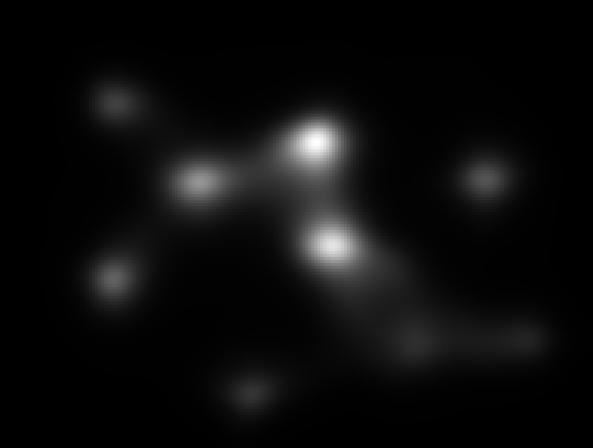}
\end{subfigure}
\begin{subfigure}[b]{0.15\linewidth}
\includegraphics[width=\linewidth]{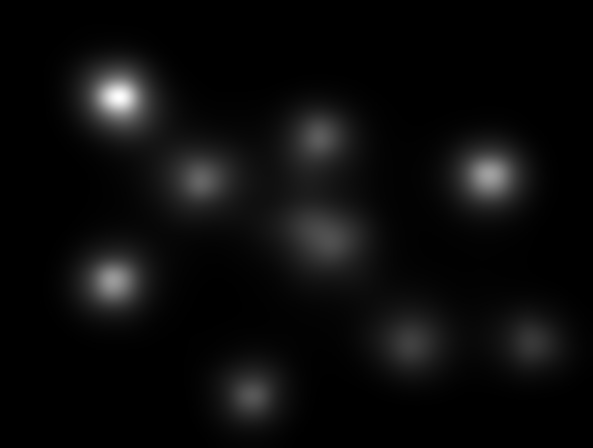}
\end{subfigure}
\begin{subfigure}[b]{0.15\linewidth}
\includegraphics[width=\linewidth]{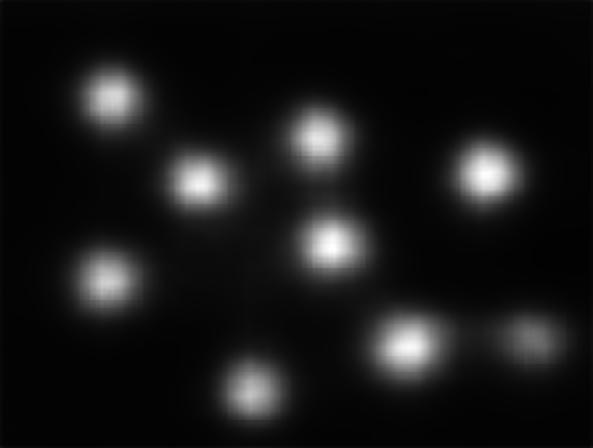}
\end{subfigure}
\begin{subfigure}[b]{0.15\linewidth}
\includegraphics[width=\linewidth]{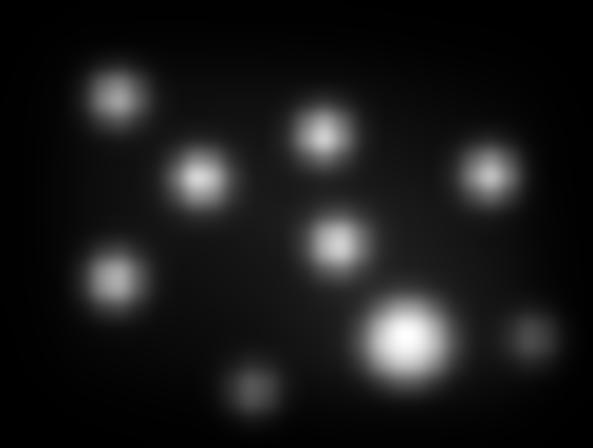}
\end{subfigure}
\begin{subfigure}[b]{0.15\linewidth}
\includegraphics[width=\linewidth]{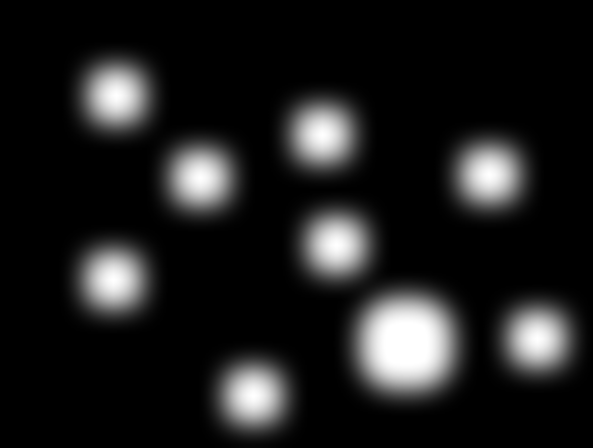}
\end{subfigure}
\caption{Some qualitative examples for deep and classical models on synthetic images, from left to right: image, predictions from DeepGaze \Romannum{2}, SAM, DVA, GBVS, and BMS.}
\label{fig:syn_com}
\end{figure}
Table~\ref{tab:dvsc} shows the performance of deep and classic models on the synthetic images. Surprisingly, although deep models achieve state-of-the-art performance on the MIT300 benchmark, they completely fail on synthetic images and are outclassed by classical models, moreover, despite the DVA~\cite{wang2018deep} has connections between shallow layer and output layer in their model, their performance is still not good on synthetic images. Fig.~\ref{fig:syn_com} illustrates saliency predictions on example stimuli by the considered methods. To investigate the reason behind this failure, we looked into the effect of fine-tuning on the inner representations and neuron responses to synthetic patterns. The results in Table~\ref{tab:syn_pre} show that deep models do indeed capture the salient patterns within the \textit{middle layers} of the architecture (\eg conv4-1 layer). Some examples (including curvature, orientation, etc) are shown in Fig.~\ref{fig:syn_actm}. Nevertheless, as indicated in Table~\ref{tab:syn_aft}, no matter how many layers are fine-tuned, the output of the deep saliency model never highlights such salient patterns. One possible reason might be that the current large databases, \eg SALICON, are biased towards natural scenes containing daily objects (text, faces, animals, cars, etc) and do not include any image similar to synthetic patters. The models, thus, do not learn anything about simple pop-out.
\begin{table}[h!]

\caption{The mean NMM for top 10 activation maps in each layer (from conv4-1 to conv5-3) in the pre-trained model for synthetic images.} 
\centering
\setlength{\tabcolsep}{3pt} 

\begin{tabular}{|c|c|c|c|c|c|}

\hline
conv5-3  & conv5-2  & conv5-1  & conv4-3  & conv4-2  & conv4-1\\
\hline
\hline
0.38  &  0.94  &  1.47  &  1.96  &  1.94  &  2.12\\
\hline
\end{tabular}
\label{tab:syn_pre}
\end{table}

\begin{table}[t!]
\caption{The mean NMM score for top 10 activation maps in the last convolutional layer for synthetic images, when fine-tuning different number of layers.}
\centering
\begin{tabular}{|c |c | c| c| c| c| c| c|}
\hline
\multicolumn{8}{|c|}{\# layers fine-tuned}\\
\hline
\hline
0 &1 &2 &3 &4 &5 &6 &all\\
\hline
0.38  & 0.8  & 0.98  &  0.92  & 0.65  & 0.86  & 0.83  & 0.73\\
\hline
\end{tabular}
\label{tab:syn_aft}
\end{table}

\section{The Influence of Task on the Learned Representations}
\begin{figure}[h!]
\centering
\includegraphics[width=0.48\textwidth]{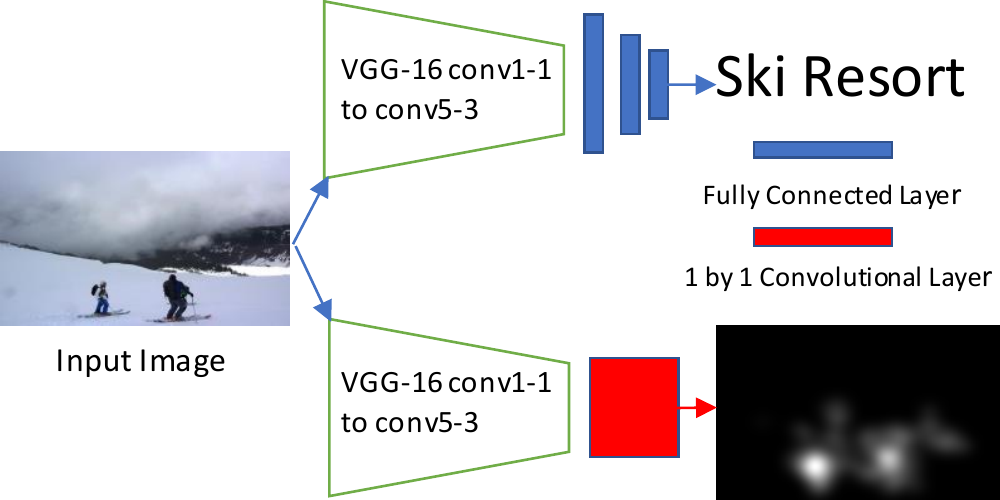}
\caption{Our architecture to study task-dependency of representations (saliency prediction vs. scene classification). We use the same data to fine-tune the pre-trained model for different tasks.}
\label{fig:sal_sce_fra}
\end{figure}
What is the driving cause for the observed change in representations after fine-tuning in previous sections? Is it due to the network being fine-tuned to a new task (saliency prediction) or is it the network being fine-tuned to a different set of data (images from a saliency prediction dataset)?
To figure out, we compare two tasks of saliency prediction and scene recognition on SALICON-SAL-SCE, which provides saliency information and scene type labels. Note that in both cases, the images used for fine-tuning are the same, therefore if the observed shift in representation is only due to the data, the inner representations should be similar in both tasks; conversely, if the task is what is driving the change, the representations should differ. 

We check three CNN trunks, including, 1) a pre-traind CNN based on VGG network for scene recognition (pt), 2) a CNN fine-tuned for saliency prediction (sp), and
3) a CNN fine-tuned for scene recognition (sr). The saliency prediction is the same as explained above. The scene recognition network consists of the VGG and 3 fully connected layers (see Fig.~\ref{fig:sal_sce_fra}). Due to the data limitation, we only fine-tuned 1 layer of the pre-trained model (layer conv5-3) for both tasks, weight balance was used when fine-tuning the scene recognition model to compensate imbalanced categories. 
\begin{table}[h!]
\setlength{\tabcolsep}{2.1pt} 
\renewcommand{\arraystretch}{1.3} 

\caption{Inner representation for different tasks (saliency prediction and scene recognition) before and after fine-tuning (pt:
pre-traind CNN based on VGG network scene recognition, sp: CNN fine-tuned for saliency prediction, sr: CNN fine-tuned for scene recognition).}
\footnotesize
\centering
\begin{tabular}{|c||c|c|c|c|c|c|c|c|c|c|c|c|}
\hline
&
\multirow{5}{*}{ \rotatebox[origin=c]{90}{ \textit{person head}}}&
\multirow{5}{*}{ \rotatebox[origin=c]{90}{ \textit{person part}} }&
\multirow{5}{*}{\rotatebox[origin=c]{90}{ \textit{animal head}} }&
\multirow{5}{*}{\rotatebox[origin=c]{90}{ \textit{animal part}} }&
\multirow{5}{*}{\rotatebox[origin=c]{90}{ \textit{object}} }&
\multirow{5}{*}{\rotatebox[origin=c]{90}{ \textit{text}}}&
\multirow{5}{*}{\rotatebox[origin=c]{90}{ \textit{symbol}}}&
\multirow{5}{*}{\rotatebox[origin=c]{90}{ \textit{vehicle}}} &
\multirow{5}{*}{\rotatebox[origin=c]{90}{ \textit{food}}} &
\multirow{5}{*}{\rotatebox[origin=c]{90}{ \textit{plant}}} &
\multirow{5}{*}{\rotatebox[origin=c]{90}{ \textit{drink}}}&
\multirow{5}{*}{\rotatebox[origin=c]{90}{ \textit{other}}} \\
& & & & & & & & & & & &\\
& & & & & & & & & & & &\\
& & & & & & & & & & & &\\
& & & & & & & & & & & &\\
\hline
\hline
&\multicolumn{12}{c|}{mean NSS for top 10 activation maps}\\
\cline{2-13}
\hline
pt  &1.21 &0.98 &2.04 &1.6 &0.69 &1.1 &1.07 &1.53 &1.25 &1.21 &1.48 &0.57    \\ 
sp  &2.51&1.6 &2.32 &1.7 &1.3 &1.44 &1.74 &2.29 &1.86 &1.58 &2.01 & 1.13   \\
sr  &0.47 &0.57 &1.54 &1.24 &0.57 &0.83 &0.61 &0.88 &0.91 &0.97 &1.28 &0.41    \\
\hline
\hline
&\multicolumn{12}{c|}{\# activation maps above threshold ($T=1.5$)}\\
\hline
pt  &2 &1 &21 &7 &0 &2 &1 &4 &1 &2 &4 &0    \\ 
sp  &44&6 &27 &10 &1 &2 &10 &19 &16 &7 &1 & 0   \\
sr  &0 &0 &4 &1 &0 &0 &0 &0 &0 &2 &3 &0    \\
\hline
\end{tabular}
\label{tab:sal_sce}
\end{table}

\begin{figure}[h!]
\centering
\begin{subfigure}[b]{0.19\linewidth}
\includegraphics[width=\linewidth]{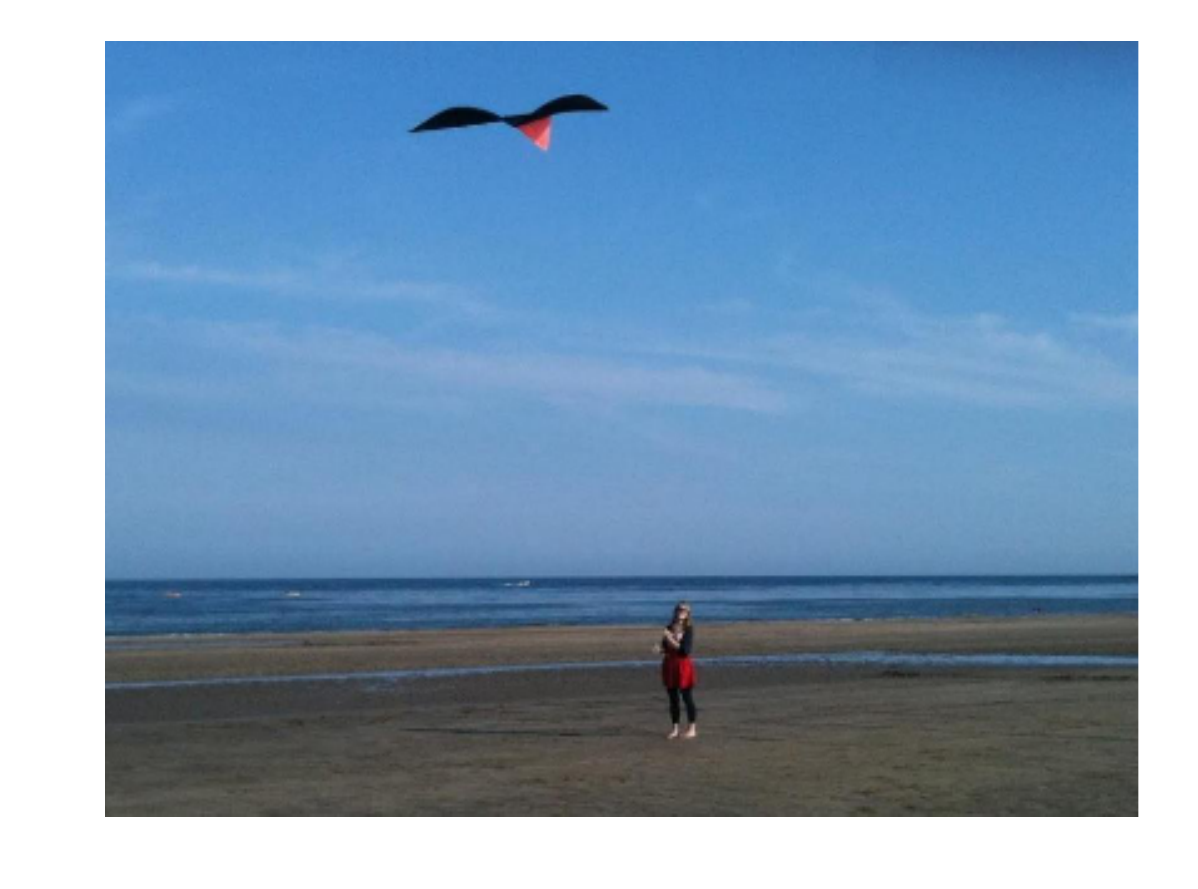}
\end{subfigure}
\begin{subfigure}[b]{0.19\linewidth}
\includegraphics[width=\linewidth]{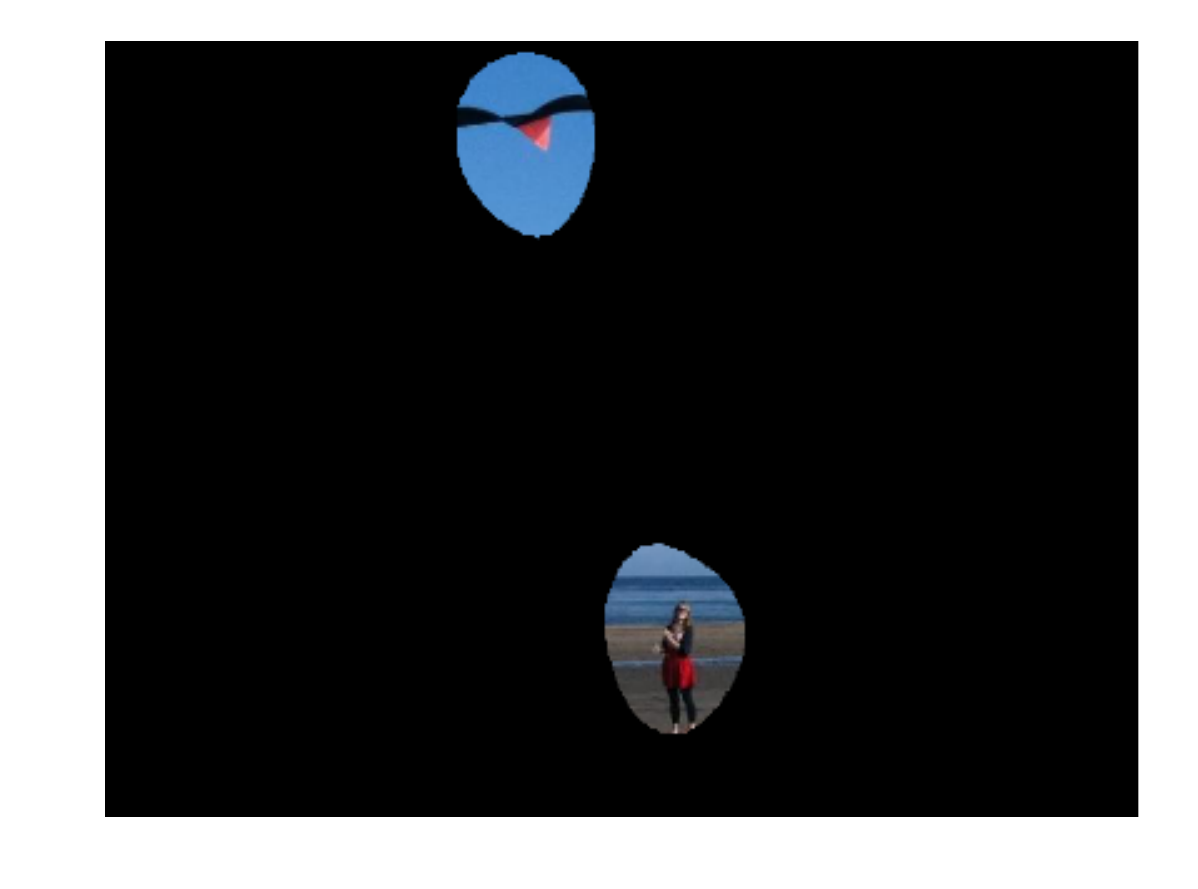}
\end{subfigure}
\begin{subfigure}[b]{0.19\linewidth}
\includegraphics[width=\linewidth]{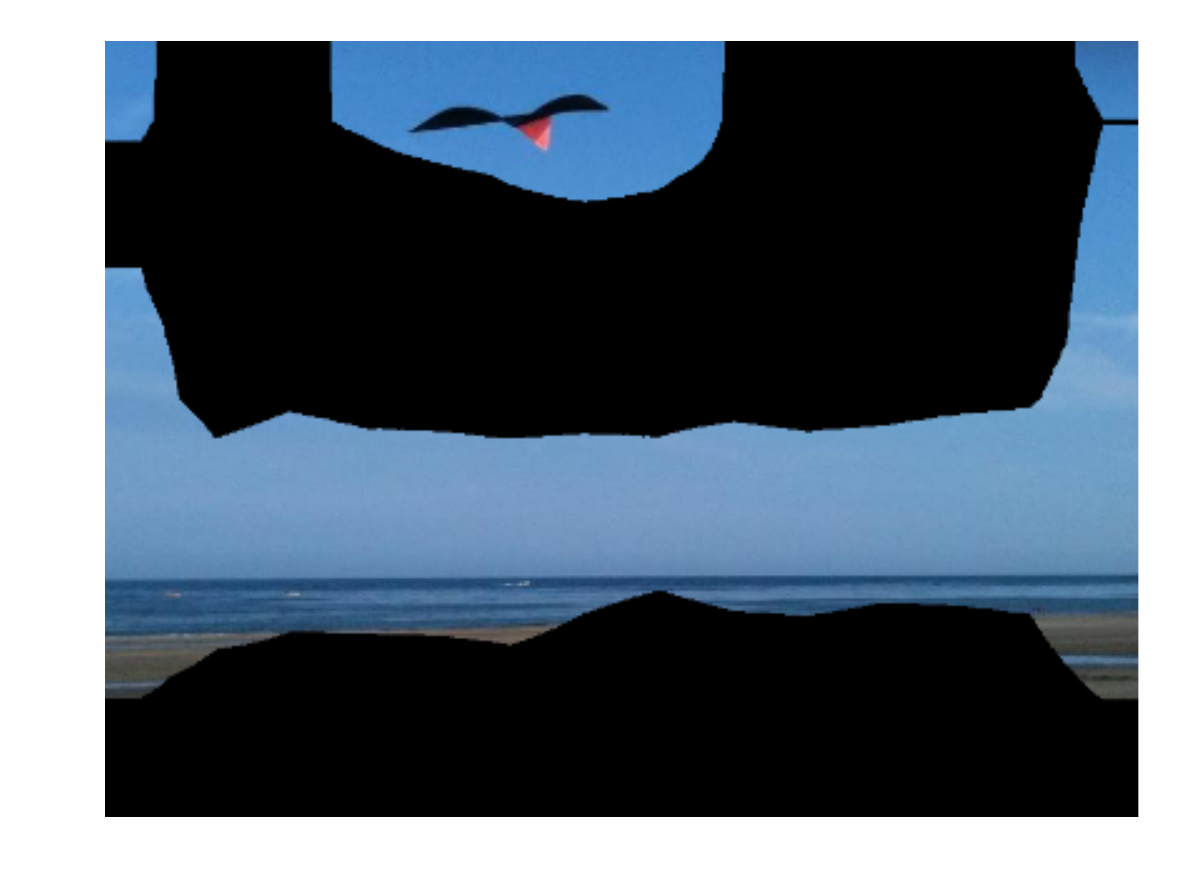}
\end{subfigure}
\begin{subfigure}[b]{0.19\linewidth}
\includegraphics[width=\linewidth]{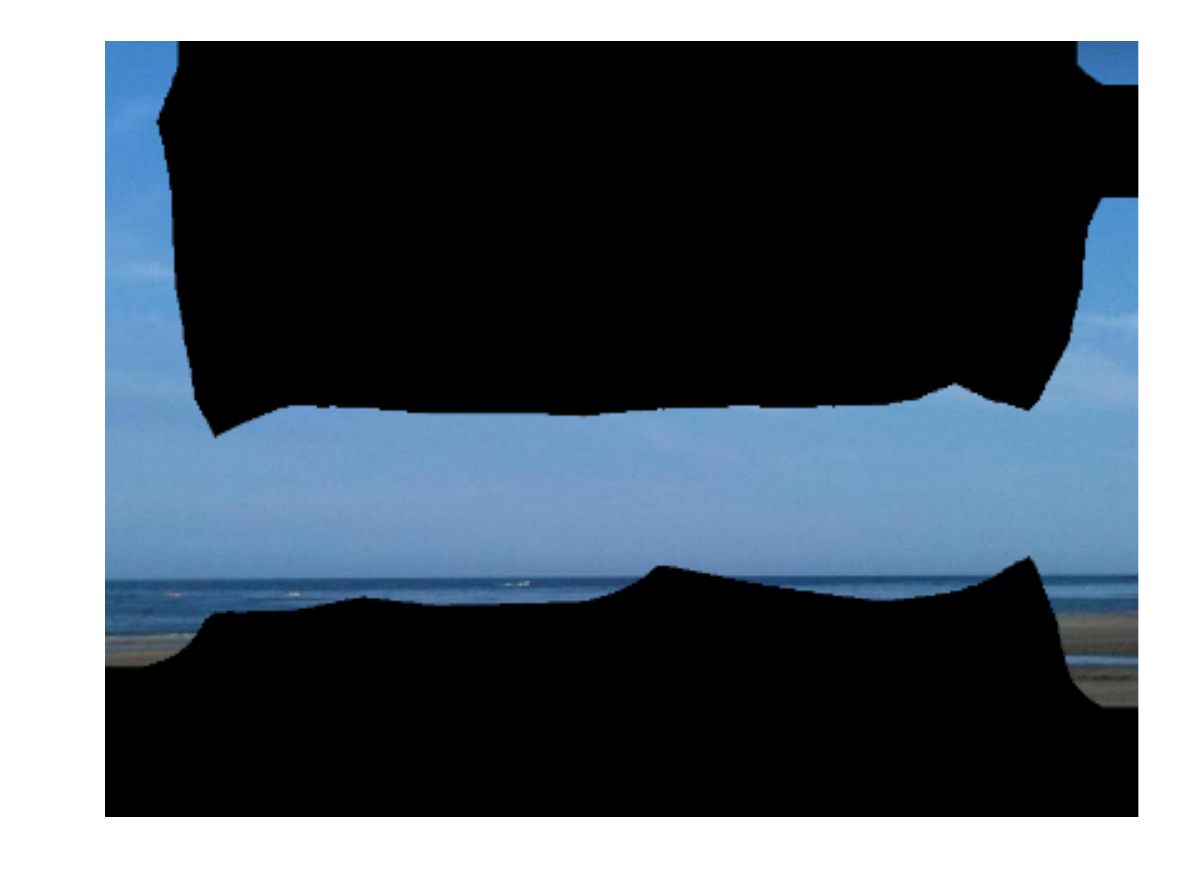}
\end{subfigure}
\begin{subfigure}[b]{0.19\linewidth}
\includegraphics[width=\linewidth]{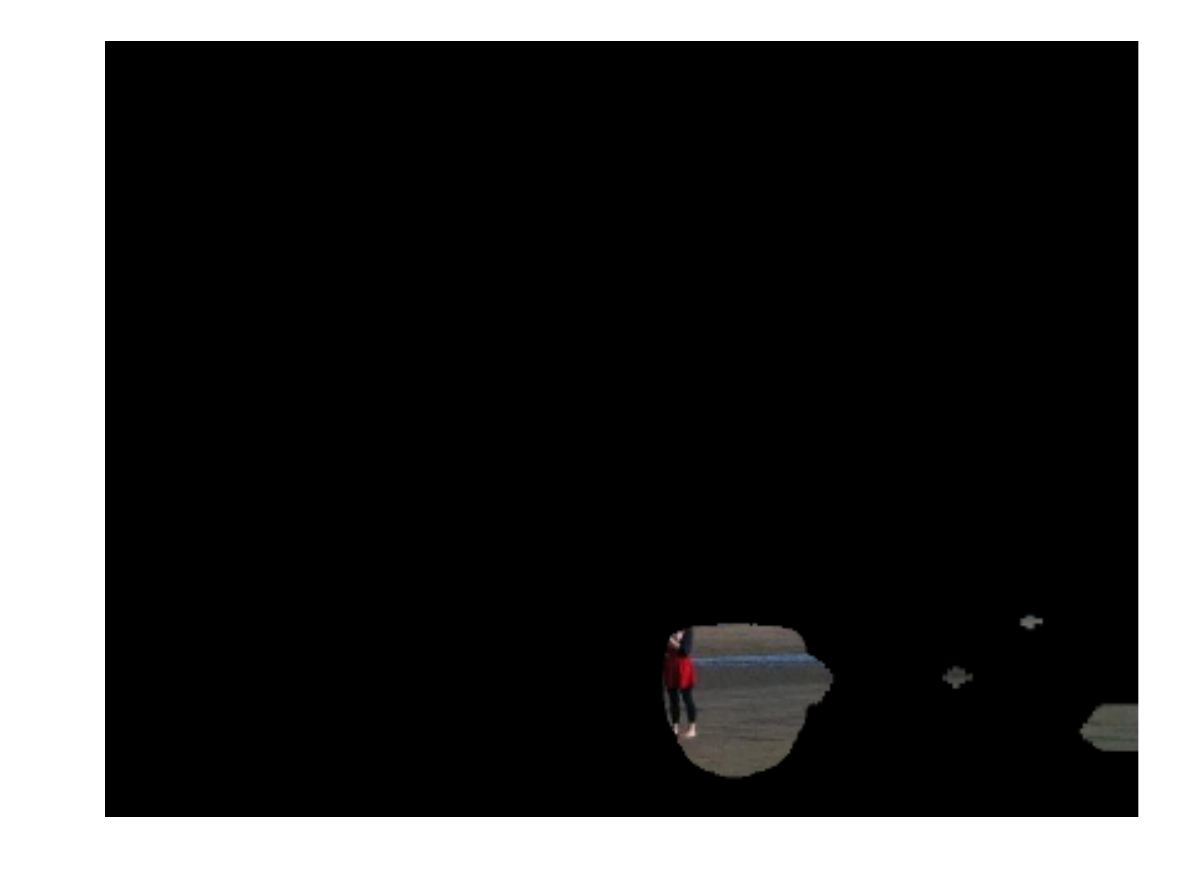}
\end{subfigure}
\begin{subfigure}[b]{0.19\linewidth}
\includegraphics[width=\linewidth]{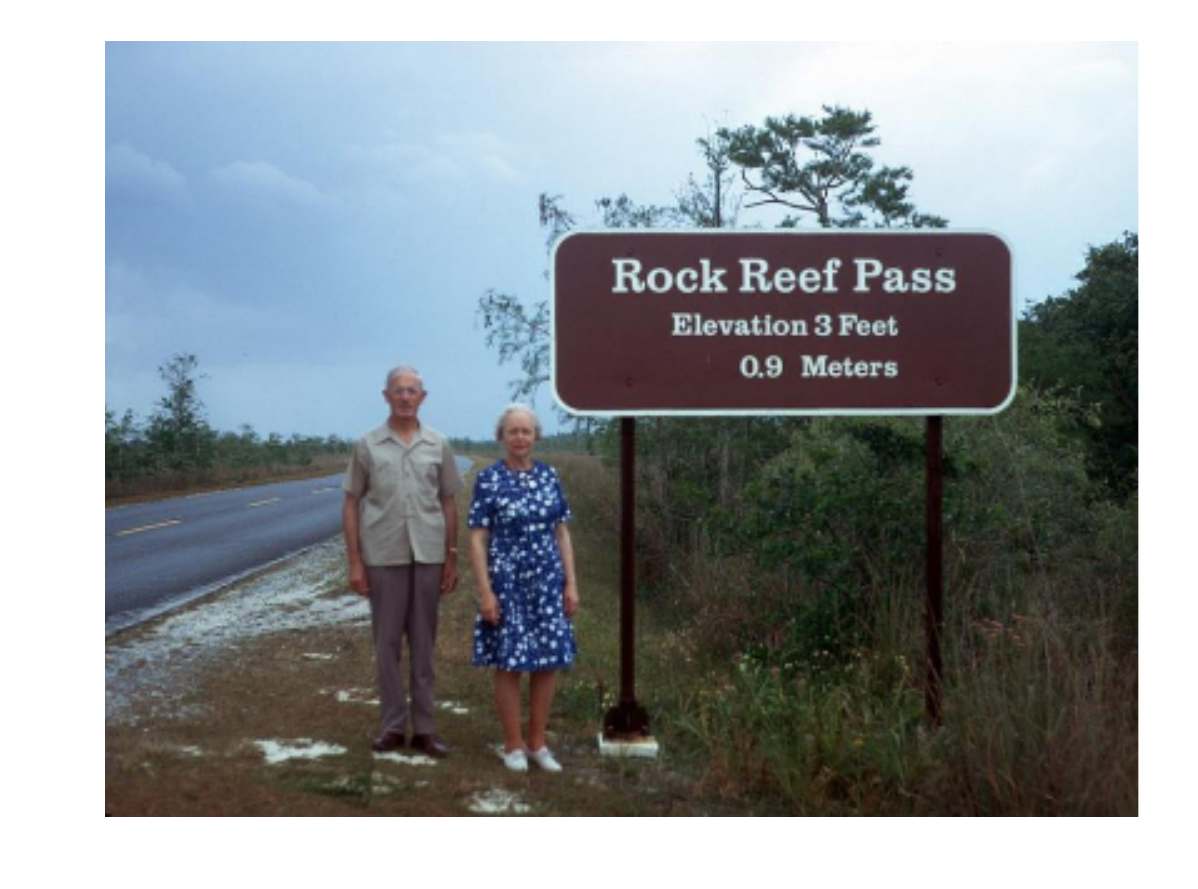}
\end{subfigure}
\begin{subfigure}[b]{0.19\linewidth}
\includegraphics[width=\linewidth]{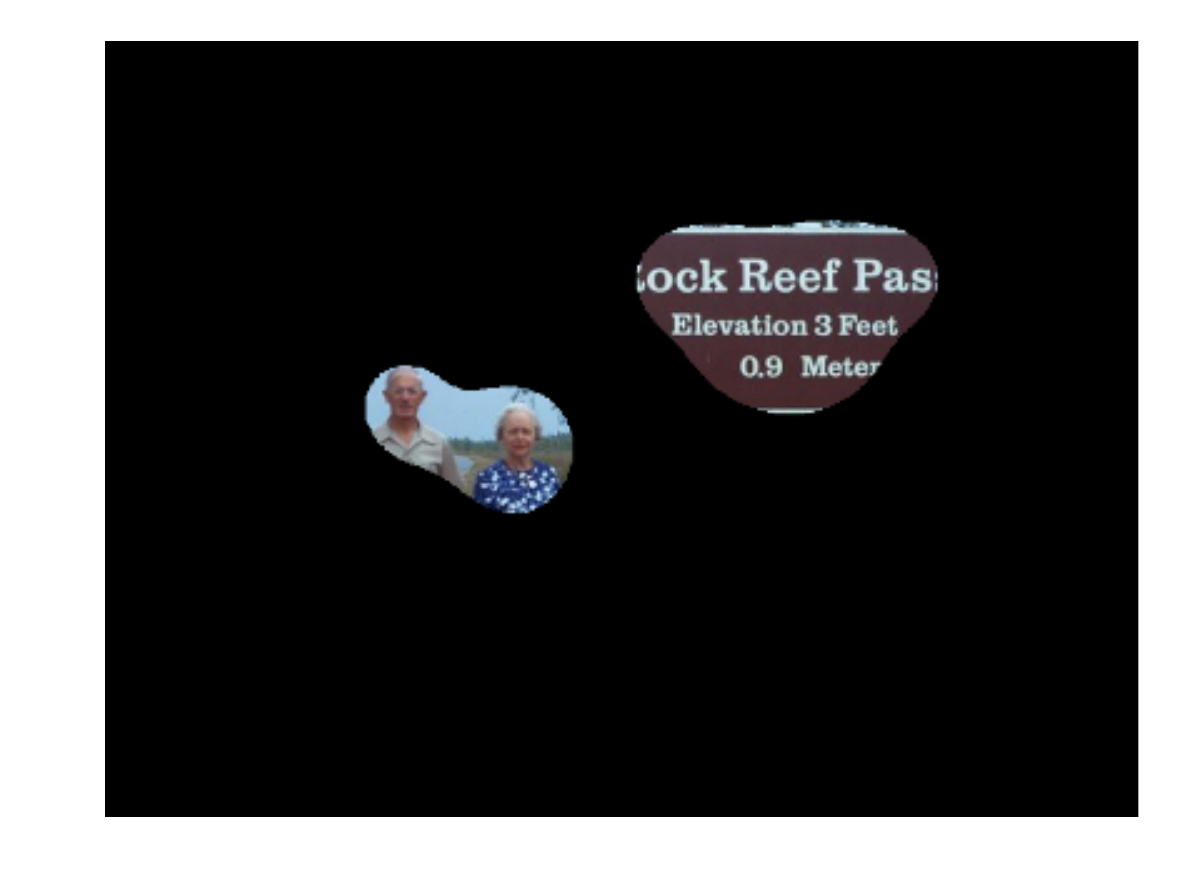}
\end{subfigure}
\begin{subfigure}[b]{0.19\linewidth}
\includegraphics[width=\linewidth]{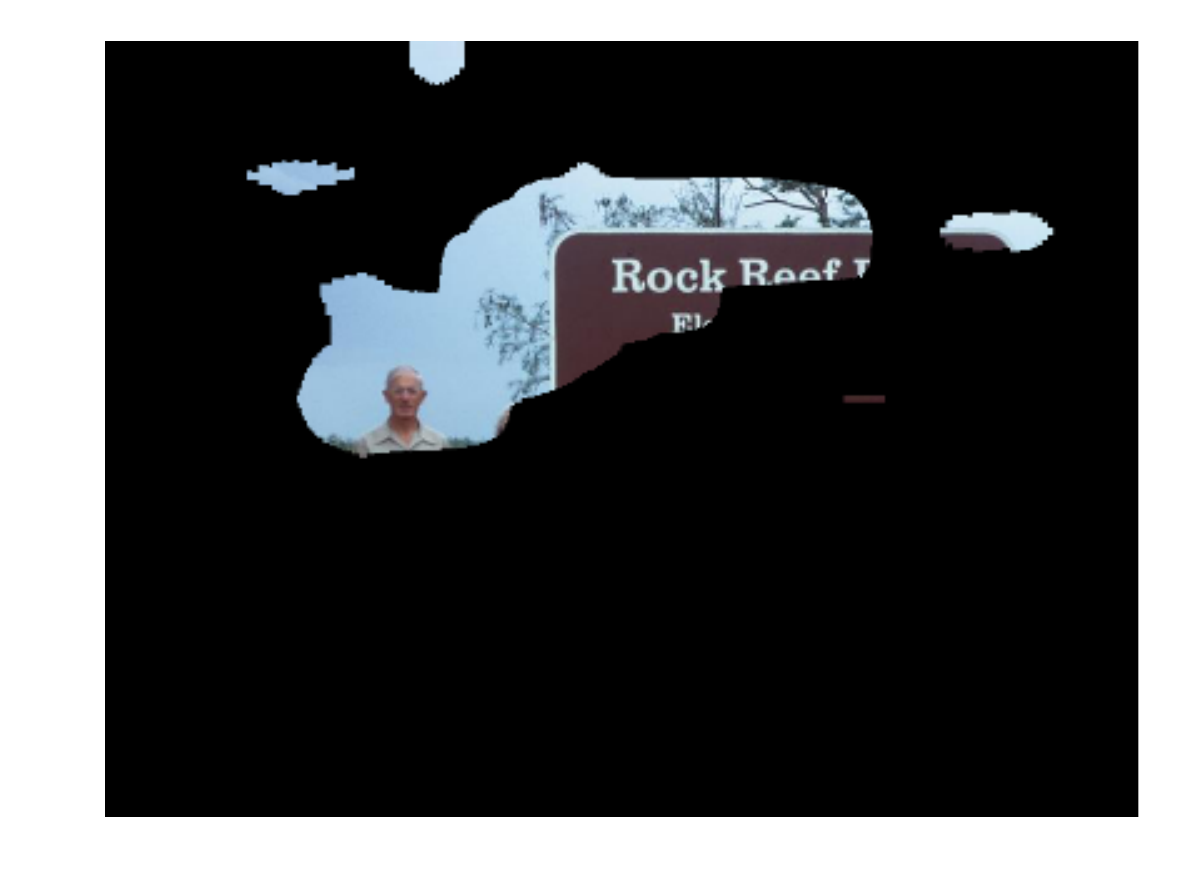}
\end{subfigure}
\begin{subfigure}[b]{0.19\linewidth}
\includegraphics[width=\linewidth]{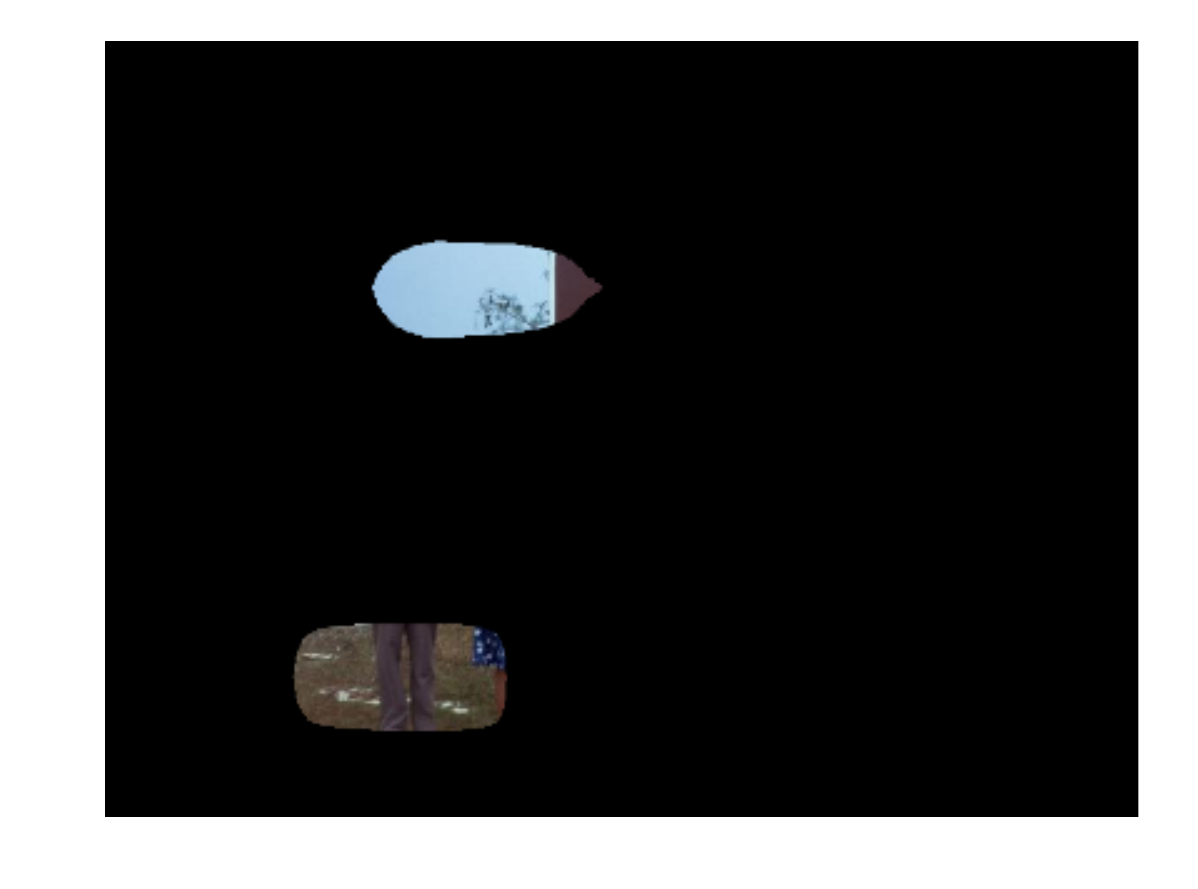}
\end{subfigure}
\begin{subfigure}[b]{0.19\linewidth}
\includegraphics[width=\linewidth]{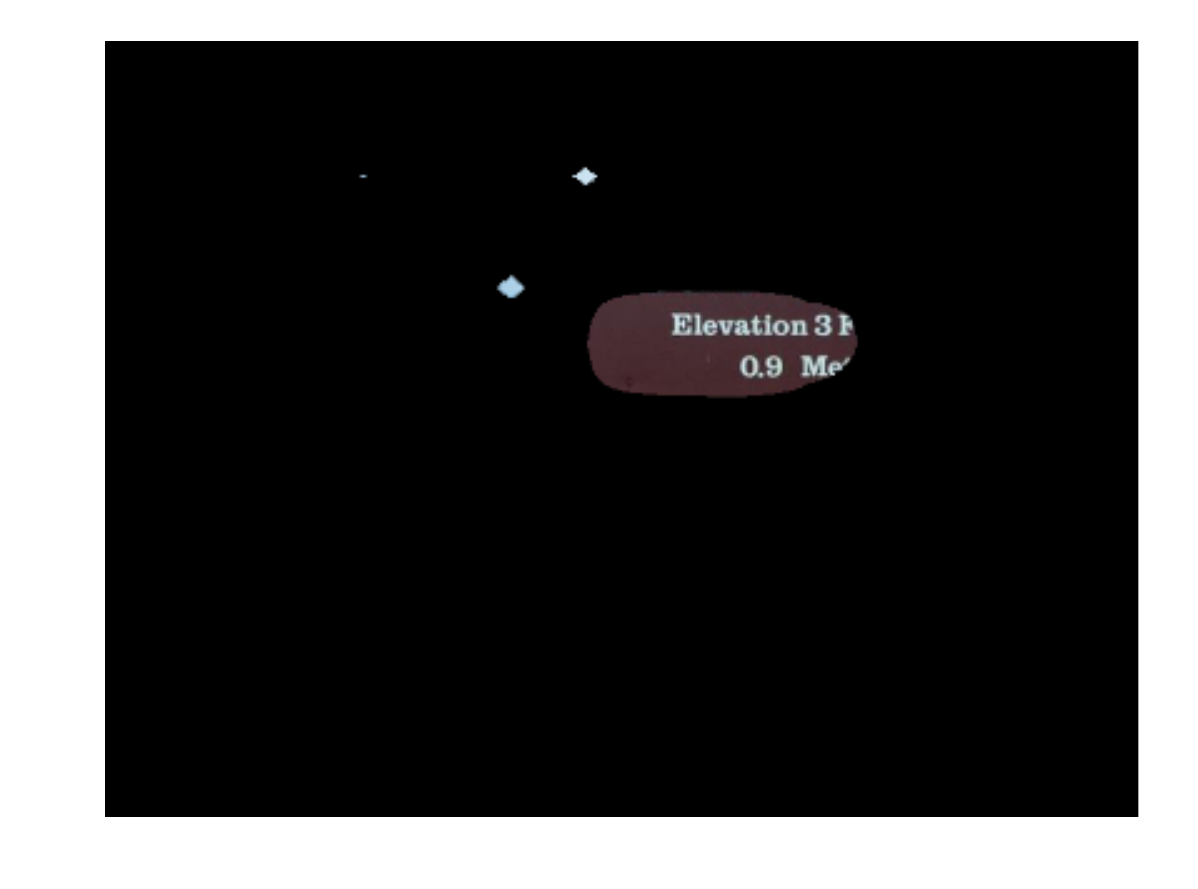}
\end{subfigure}
\caption{From left to right: Original image, image overlapped with the ground-truth fixation map, overlapped with the activation map by the pre-trained model, overlapped with the activation map after fine-tuning for scene recognition, overlapped with the activation map after fine-tuning for saliency prediction.}
\label{fig:sal_sce_f}
\end{figure}

Table~\ref{tab:sal_sce} shows how the inner representation changes with respect to saliency categories, once the CNN is fine-tuned for different tasks using the same data. The results show that fine-tuning for saliency prediction drives the inner representations to became more selective to salient categories, while fine-tuning for scene recognition leads to less selectivity to salient categories and inhibition of some other salient regions. Examples of the activation map change for each task is provided in Fig.~\ref{fig:sal_sce_f}.

\begin{table}[h!]
\caption{The NSS scores of mean activation maps for correct and wrong prediction in scene recognition task.}
\centering
\begin{tabular}{|c|c|}
\hline
correct prediction&wrong prediction\\
\hline
\hline
0.12&0.11\\
\hline
\end{tabular}
\label{tab:sce_hu}
\end{table}

To what degree the inner representations in the scene recognition network is consistent with human attention? Does the scene recognition model attend to the locations a human may find salient? We investigate this by computing the NSS score between the attention of model (the mean of 512 activation maps in layer conv5-3) and the human fixation on the image. The results are summarized in Table~\ref{tab:sce_hu}, showing that
the NSS score is small irrespective of whether the model's prediction is correct or not. In other words, the model's attention in scene recognition is different from human attention in free-viewing. 

To summarize, the above results indicate that the inner representations, during fine-tuning the same CNN for the two different tasks of saliency prediction and scene recognition on the same data, mostly change because of the task and not the data.

\section{Discussion and Conclusion}

In this work, we analyzed the internals of deep saliency models.
To this end, we annotated 3 datasets and conducted several experiments to unveil the secrets of deep saliency models. Our analysis on this data revealed that a deep neural network pre-trained for image recognition already encodes some visual saliency in the image. Fine-tuning this pre-trained model for saliency prediction produces a model with uneven response to saliency categories, \eg neurons sensitive to textual input start attending more to human head.
We showed that although deep models do capture synthetic pop-out stimuli within their inner layers, they fail to predict such salient patterns in their output, contrary to classical models of saliency prediction.
We also confirmed that the observed change in the inner representations after fine-tuning 
is mainly due to fine-tuning for the task and not the data. In our study, fine-tuning the model for saliency prediction resulted in more selective responses to salient regions, though uneven. On the other hand, fine-tuning the model for scene recognition had inhibitory effect and the inner representations were losing their selectivity to some of the existing salient patterns.

To conclude, pushing the development of better data-driven deep visual saliency models further may require delicate attention to the diversity of salient categories within images. In other words, we may need not only a large scale dataset, but also a dataset with rich saliency categories to ensure generalization.

\vspace{-5pt}
\section*{Acknowledgements}
This research was supported by the EPSRC project DEVA EP/N035399/1. Dr Pugeault acknowledges funding from the Alan Turing Institute (EP/N510129/1). H. R. Tavakoli acknowledges NVIDIA for the donation of GPUs used in his research.
{\small
\bibliographystyle{ieee}
\bibliography{egbib}
}

\end{document}